\documentclass[review]{elsarticle}
\usepackage{graphicx,amssymb,mathrsfs,amsmath,pifont,amscd,latexsym,amsfonts,amsthm,colortbl,bm}
\usepackage{algpseudocode}
\usepackage[linesnumbered,ruled]{algorithm2e}
\usepackage{geometry}
\usepackage{graphics}
\usepackage{epstopdf}
\usepackage{float}
\usepackage{subfigure}
\usepackage{epsfig}
\usepackage{setspace}
\usepackage{booktabs}
\usepackage{xcolor}
\usepackage{multirow}
\usepackage{lineno}
\usepackage{color}
\usepackage[colorlinks,urlcolor=blue,driverfallback=dvipdfm]{hyperref}

\newcommand{\cmark}{\ding{51}}
\newcommand{\xmark}{\ding{55}}
\graphicspath{{figures/}}

\begin{document}
\begin{frontmatter}

\title{Cross-City Matters: A Multimodal Remote Sensing Benchmark Dataset for Cross-City Semantic Segmentation using High-Resolution Domain Adaptation Networks}

\author[secondaddress]{Danfeng Hong}
\ead{hongdf@aircas.ac.cn}

\author[secondaddress,thirdaddress]{Bing Zhang\corref{correspondingauthor}}
\cortext[correspondingauthor]{Corresponding author}
\ead{zb@radi.ac.cn}

\author[fourthaddress]{Hao Li}
\ead{hao_bgd.li@tum.de}

\author[secondaddress]{Yuxuan Li}
\ead{liyuxuan231@mails.ucas.ac.cn}

\author[secondaddress]{Jing Yao}
\ead{yaojing@aircas.ac.cn}

\author[fifthaddress]{Chenyu Li}
\ead{lichenyu@seu.edu.cn}

\author[fourthaddress]{Martin Werner}
\ead{martin.werner@tum.de}

\author[sixthaddress,secondaddress]{Jocelyn Chanussot}
\ead{jocelyn.chanussot@grenoble-inp.fr}

\author[seventhaddress]{Alexander Zipf}
\ead{zipf@uni-heidelberg.de}

\author[eighthaddress]{Xiao Xiang Zhu}
\ead{xiaoxiang.zhu@tum.de}

\address[secondaddress]{Aerospace Information Research Institute, Chinese Academy of Sciences, Beijing 100094, China;}
\address[thirdaddress]{College of Resources and Environment, University of Chinese Academy of Sciences, Beijing 100049, China;}
\address[fourthaddress]{Big Geospatial Data Management, Technical University of Munich, Munich 85521, Germany;}
\address[fifthaddress]{School of Mathematics, Southeast University, Nanjing 210096, China;}
\address[sixthaddress]{Univ. Grenoble Alpes, CNRS, Grenoble INP, GIPSA-Lab, Grenoble 38000, France;}
\address[seventhaddress]{GIScience Chair, Institute of Geography, Heidelberg University, Heidelberg 69120, Germany;}
\address[eighthaddress]{Data Science in Earth Observation, Technical University of Munich, Munich 80333, Germany.}

\begin{abstract}
 Artificial intelligence (AI) approaches nowadays have gained remarkable success in single-modality-dominated remote sensing (RS) applications, especially with an emphasis on individual urban environments (e.g., single cities or regions). Yet these AI models tend to meet the performance bottleneck in the case studies across cities or regions, due to the lack of diverse RS information and cutting-edge solutions with high generalization ability. To this end, we build a new set of multimodal remote sensing benchmark datasets (including hyperspectral, multispectral, SAR) for the study purpose of the cross-city semantic segmentation task (called C2Seg dataset), which consists of two cross-city scenes, i.e., Berlin-Augsburg (in Germany) and Beijing-Wuhan (in China). Beyond the single city, we propose a \textbf{high}-resolution \textbf{d}omain \textbf{a}daptation \textbf{n}etwork, HighDAN for short, to promote the AI model's generalization ability from the multi-city environments. HighDAN is capable of retaining the spatially topological structure of the studied urban scene well in a parallel high-to-low resolution fusion fashion but also closing the gap derived from enormous differences of RS image representations between different cities by means of adversarial learning. In addition, the Dice loss is considered in HighDAN to alleviate the class imbalance issue caused by factors across cities. Extensive experiments conducted on the C2Seg dataset show the superiority of our HighDAN in terms of segmentation performance and generalization ability, compared to state-of-the-art competitors. The C2Seg dataset and the semantic segmentation toolbox (involving the proposed HighDAN) will be available publicly at \url{https://github.com/danfenghong}.
\end{abstract}

\begin{keyword} 
Cross-city, Deep Learning, Dice Loss, Domain Adaptation, High-resolution Network, Land Cover, Multimodal Benchmark Datasets, Remote Sensing, Segmentation.
\end{keyword}

\end{frontmatter}
\section{Introduction}

Remote sensing (RS) is an essential means to acquire large-scale and high-quality Earth observation (EO) data in a concise time, which significantly advances the development of EO techniques. However, the conventional expert system-centric mode has run into bottlenecks and cannot meet the EO demand of the RS big data era well, particularly when facing complex urban scenes. Artificial Intelligence (AI) techniques provide one viable option that is capable of finding out potentially valuable knowledge from the vast amounts of pluralistic EO data more intelligently, enabling the understanding and monitoring of the contemporary urban environment.

These advanced AI models, e.g., deep learning, have been successfully applied for various RS and geoscience applications, which have been proven to be particularly applicable to the unitary urban environment where the types, characteristics, and spatial distributions of surface elements are significantly consistent and similar. Nevertheless, the ability to address multiple urban environmental issues with highly spatio-temporal and regional change remains limited. The possibly feasible solutions are two-fold. On the one hand, the joint exploitation of multimodal RS data has been proven to be helpful to improve the processing ability of cross-city or regional cases, since the RS data acquired from different platforms or sensors can provide richer and more diverse complementary information. On the other hand, designing more leading-edge AI models with a focus on promoting the generalization ability across cities or regions is an inexorable trend to alleviate the semantic gap between different urban environments, making it mutually transferable for knowledge.

In recent years, enormous efforts have been made to couple or jointly analyze different RS observation sources by the attempts to design advanced fusion and interpretation methods to achieve a more diversified description for the studied urban scene. In particular, a growing body of studies has confirmed the achievement of multimodal AI models in one single urban environment. It should be noted, however, that multi-city-related cases are evolving at a relatively slow speed. This slow progression can be satisfactorily explained by two very likely reasons as follows.
\begin{itemize}
    \item One refers to the lack of high-quality multimodal RS benchmark datasets for a better understanding of cross-city environments.
    \item Another is that currently developed methodologies prefer to focus on extreme performance pursuit in one single urban environment rather than improve the model generalization ability, particularly for diverse urban environments (e.g., different cities or regions).
\end{itemize}

To boost technical breakthroughs and accelerate the development of EO applications across cities or regions, creating a multimodal RS benchmark dataset for cross-city land cover segmentation makes necessary. Just as important, the high generalization ability in terms of methodology development is of paramount importance. This drives us to develop such a model with high transferability between different cities or regions by means of domain adaption (DA) techniques. Numerous experiments will be conducted on the cross-city land cover segmentation dataset to show the superiority of DA-based approaches over those semantic segmentation algorithms that do not consider knowledge transfer across domains. More specifically, our contributions in this paper can be unfolded as follows. 
\begin{itemize}
    \item A new set of multimodal RS benchmark datasets is built for the study purpose of the cross-city semantic segmentation task, named C2Seg for short. C2Seg consists of two subsets, i.e., Berlin-Augsburg (in Germany) dataset collected from EnMAP, Sentinel-2, and Sentinel-1, respectively, Beijing-Wuhan (in China) dataset collected from Gaofen-5, Gaofen-6, and Gaofen-3, respectively. The C2Seg dataset will be available freely and publicly, promoting the research progress on semantic segmentation across cities or regions substantially. To the best of our knowledge, C2Seg is the first benchmark dataset about the cross-city multimodal RS image segmentation task, which considers the three-modality study case, including hyperspectral, multispectral, and synthetic aperture radar (SAR) data acquired from the currently well-known satellite missions. {The C2Seg datasets have been utilized for the WHISPERS2023 conference \url{https://www.ieee-whispers.com/} in the capacity of Challenge 1: Cross-City Multimodal Semantic Segmentation. These datasets are accessible at \url{https://www.ieee-whispers.com/cross-city-challenge/}, with the training data already made available. Shortly, we plan to make all datasets, including both training and testing data, accessible to the wider research community.}
    \item A high-resolution domain adaptation network (HighDAN) is devised to bridge the gap between RS images from different urban environments utilizing adversarial learning, thereby making it possible to transfer the learned knowledge from one domain to another effectively and eliminate inter-class variations to a great extent. Further, HighDAN, which is built based on the high-resolution network (HR-Net), is capable of capturing multi-scaled image representations from parallel high-to-low-resolution subnetworks, yielding repetitive information exchange across different resolutions in a highly efficient manner.
    \item To reduce the impact of the sample number imbalance between classes due to the multi-city studies, the Dice loss is considered and embedded in the proposed HighDAN.
\end{itemize}
 
The remaining sections of the paper are organized as follows. Section 2 reviews the related work for semantic segmentation in the land cover classification task systematically from the perspectives of individual study scenes and cross-region (or cross-city) cases. Section 3 introduces the newly-built datasets and correspondingly elaborates on the proposed methodology. Experiments are conducted on the datasets with extensive discussion and analysis in Section 4. Finally, Section 5 makes the conclusion of this paper with some remaining challenges and plausible future solutions.

\section{Related Work}

Over the past decade, deep learning (DL) has been garnering increasing attention in many application fields \cite{yuan2020deep}, owing to its powerful ability for data representation and learning. In particular, the ever-perfecting DL techniques for RS enable accurate and automatic land cover mapping. According to different studied scenes, we divide these approaches into individual environments and multi-region (or city) ones, where single-modality and multimodal RS data are further involved.

\subsection{Semantic Segmentation on Individual Environments}

With the emergence and rapid development of DL, there have been recently numerous semantic segmentation methods successfully developed for RS with a focus on a single studied scene \cite{yuan2021review}. Kampffmeyer \textit{et al.} \cite{kampffmeyer2016semantic} developed deep convolutional neural networks (CNNs) for semantic segmentation in terms of small objects in urban areas, where the uncertainty in CNNs is modeled by Bayesian approximation in Gaussian process \cite{gal2016dropout}. The CNNs-based architecture was also used in \cite{kemker2018algorithms} for semantic segmentation on multispectral RS images rather than high-resolution RGB images. In this work, synthetic multispectral images are generated for initializing deep CNNs to alleviate the effects of label scarcity. Yi \textit{et al.} \cite{yi2019semantic} proposed a deep residual U-Net (ResUNet) framework, which consists of cascade down-sampling and up-sampling subnetworks, for urban building extraction using very high-resolution (VHR) RS images. Further, Diakogiannis \textit{et al.} \cite{diakogiannis2020resunet} designed an enhanced ResUNet version, ResUNet-a, with atrous convolutions for semantic segmentation of RS images. A multi-scale semantic segmentation network was proposed in \cite{du2021mapping} for fine-grained urban functional zone classification using VHR RS images and object-based strategies. {Adding to this advancement, Wang \textit{et al.} \cite{wang2022unetformer} introduced a recent breakthrough in the field, unveiling an efficient U-shaped transformer network custom-tailored for the precise execution of semantic segmentation tasks in VHR urban scene images. Concurrently, He and his collaborators \cite{he2022swin} incorporated the Swin transformer into the U-Net architecture, further enhancing the capabilities of semantic segmentation in RS applications. In a recent development, as documented in \cite{wang2023seismic}, a novel approach following the SegFormer \cite{xie2021segformer} framework, enriched by the utilization of hypercolumns, has been employed for seismic facies segmentation.} Although these DL approaches have provided superior segmentation accuracy over traditional model-driven models on single-modality RS images, they inevitably meet the performance bottleneck in the complex scene understanding task (due to the lack of diverse modality information).  

With the ever-growing availability of RS data sources from well-known spaceborne and airborne missions, e.g., Gaofen in China, Sentinel in the EU, and Landsat in the USA, multimodal RS techniques have been garnering increasing attention and made extraordinary progress in various EO-related tasks. The data acquired by different platforms can provide diverse and complementary information \cite{dalla2015challenges}. The joint exploitation of different RS data has been therefore proven to be effective in further enhancing our understanding, possibilities, and capabilities in a single urban environment. As the mainstream application, semantic segmentation of multimodal RS images using DL has been widely studied in recent years. Audebert \textit{et al.} \cite{audebert2016semantic} extracted the multi-scaled deep features from multimodal EO data for semantic labeling. Further, the same authors extended their work in \cite{audebert2016semantic} by implementing the multi-scale deep fully convolutional networks (FCNs) \cite{long2015fully, wu2019fastfcn} based on SegNet \cite{badrinarayanan2017segnet} to process and understand multimodal RS data for land cover segmentation \cite{audebert2018beyond}. Similar to \cite{hong2020deep}, they also discussed the fusion strategies of different RS modalities, e.g., early, middle, and late fusion. In \cite{wieland2019multi}, multi-sensor cloud and shadow segmentation are investigated using CNNs. Wurm \textit{et al.} \cite{wurm2019semantic} proposed to transfer FCNs trained from external datasets for improving the semantic segmentation performance of cross-modal satellite images. Segal \textit{et al.} \cite{segal2020cloud} designed a CNNs-based cloud detection algorithm based on the Deeplab architecture \cite{chen2017deeplab} for multimodal satellite images, achieving an effective detection performance improvement. Ren \textit{et al.} \cite{ren2022dual} proposed a dual-stream high-resolution network (HR-Net) \cite{sun2019deep} for the deep fusion of GF-2 and GF-3 multimodal RS data for land cover classification. {In the work by Adriano \textit{et al.} \cite{adriano2021learning}, the authors explored the mapping and evaluation of building damage from a segmentation perspective, leveraging the rich information provided by multimodal and multitemporal RS data, marking a significant advancement in the field of damage assessment. }

\subsection{Semantic Segmentation across Regions or Cities}

Currently developed semantic segmentation networks of RS images in terms of the design of network architecture, module details, and the use of loss functions have reached their performance peak. It is a noticeable phenomenon, however, that these models are more often than not well-designed for individual study scenes. This will lead to poor generalization ability for the model, which can not well match the level of the segmentation performance, particularly in the cases of cross-city or cross-region studies. For this reason, researchers have started gradually paying more attention to the task of semantic segmentation across regions or cities. 

Domain adaptation (DA) has been proven to be helpful in reducing the semantic gap between source and target domains \cite{ganin2015unsupervised}. The DA-related approaches have been recently designed to address the challenge of cross-scene RS image semantic segmentation. For example, Chen \textit{et al.} \cite{chen2017no} proposed a road scene adaptation segmenter by utilizing high-resolution RS images from Google Street View in an unsupervised manner, which is well-designed to solve the problem of dataset biases across different cities effectively. A novel adversarial learning method was presented in \cite{tsai2018learning} for DA in semantic segmentation, where the spatially structural similarity is employed to narrow down the gap between data distribution differences of different domains. Tong \textit{et al.} \cite{tong2020land} first pre-trained a deep CNN with a well-annotated Gaofen-2 land cover dataset, and transferred the trained deep model for the unlabeled RS image classification in the target domain. By contrast, Zhu \textit{et al.} \cite{zhu2021deep} directly learned a transfer network by attempting to align the data distribution of subdomains with the utilization of a local maximum mean discrepancy for image classification. Li \textit{et al.} \cite{li2022improving} proposed a few-shot transfer learning (FSTL) method to improve the generalization capability of pre-trained deep CNN on mapping human settlement across countries. Li \textit{et al.} \cite{li2021learning} reduced the impact of data shift effectively by designing weakly-supervised constraints, making it more suitable for the task of cross-domain RS image semantic segmentation. {Moreover, Wang \textit{et al.} \cite{wang2022cross} contributed to the field by facilitating domain adaptation (DA) in the context of cross-sensor VHR urban land cover segmentation, with a focus on accommodating both airborne and spaceborne RS images. Further, the same investigators \cite{ma2023domain} extended their work for semantic segmentation in RS by considering local consistency and global diversity to enhance the DA capability.}

The joint use of multimodal RS data is capable of better mining the representation ability of diverse RS modalities, further weakening the effects of data shift to some extent when the model is trained on one RS data domain and transferred to another. Hong \textit{et al.} \cite{hong2020x} aimed at the semi-supervised transfer learning challenge for cross-scene land cover semantic classification in RS and accordingly proposed a cross-modal deep network, called X-ModelNet. The same authors in \cite{hong2021multimodal} further extended their work with two plug-and-play adversarial modules to enhance the robustness and transferability of cross-region RS image semantic segmentation. Similarly, Ji \textit{et al.} \cite{ji2021generative} fully aligned the source and target domains in the generative adversarial network (GAN) \cite{goodfellow2020generative} guided image space. The style translation technique is utilized to train an end-to-end deep FCN with a combination of DA and semantic segmentation from the multi-source RS images to identify the different types of land cover elements. Zhao \textit{et al.} \cite{zhao2022cross} reduced the disparity across scenes by using fractional Fourier fusion and spatial-spectral DA techniques for cross-domain multi-source RS data classification. These aforementioned methods can be unified into a general multimodal deep learning framework for RS image land cover classification (i.e., MDL-RS) on both individual and cross-region environments \cite{hong2021more}.

There have recently been certain researches developed by attempts to investigate the feasibility and effectiveness of semantic segmentation across regions or cities using multimodal RS images. Yet the inadequate integration among high-performance deep semantic segmentation architectures, DA networks, and the use of multimodal RS data inevitably leads to the performance bottleneck in cross-domain land cover classification. Most importantly, the problems in the lack of multimodal RS benchmark datasets become obstacles to the development of urban RS and further decelerate the technical progress of scientific research in terms of cross-city semantic segmentation. 

The follow-up two sections will therefore focus on the solutions to the two above-mentioned difficulties. Accordingly, one creates large-scale multimodal RS benchmark datasets for the study of cross-city semantic segmentation and another brings forth new ideas in the update and upgrade of network architecture and blending between multimodal RS data and DA techniques.

\section{C2Seg: A Multimodal RS Dataset for Cross-City Semantic Segmentation}
\subsection{Overview}
To overcome the difficulty of multimodal RS data shortage and boost the technological innovation of urban scene understanding across cities, we build a new collection of multimodal RS benchmark datasets, including hyperspectral, multispectral, and SAR data, for research into cross-city semantic segmentation (i.e., C2Seg). C2Seg datasets consist of two cross-city scenes as follows.
\begin{itemize}
    \item C2Seg-AB: Berlin-Augsburg cities in Germany, which are collected from EnMAP, Sentinel-2, and Sentinel-1 satellite missions on the date as close as possible, and accordingly pre-processed via ESA's SNAP toolbox.
    \item C2Seg-BW: Beijing-Wuhan cities in China, which are collected from Gaofen-5, Gaofen-6, and Gaofen-3 satellite missions on the date as close as possible, and pre-processed using the ENVI software.
\end{itemize}

{In contrast to certain well-known HR or VHR datasets, such as OpenEarthMap \cite{xia2023openearthmap}, it's worth noting that our C2Seg datasets encompass three distinct RS modalities, even though they maintain a GSD of only 10 meters. Furthermore, we are committed to fostering research progress in the domain of cross-city semantic segmentation by making the C2Seg datasets openly available for free download. These datasets encompass 13 distinct land use and land cover semantic categories\footnote{They are \textit{Urban Fabric}, \textit{Industrial/Commercial/Transport Units}, \textit{Mine/Dump/Construction Sites}, \textit{Artificial/Non-Agricultural/Vegetated Areas}, \textit{Surface Water}, \textit{Street}, \textit{Arable Land}, \textit{Permanent Crops}, \textit{Pastures}, \textit{Forests}, \textit{Shrub and/or Herbaceous Vegetation Associations}, \textit{Open Spaces with Little or Non-Vegetation}, and \textit{Inland Wetlands}.}. To the best of our knowledge, this represents a pioneering effort in creating a large-scale benchmark dataset tailored for cross-city multimodal RS semantic segmentation, taking into account three kinds of RS modalities. The C2Seg datasets will be unfolded in detail as follows.}

\begin{figure*}[!t]
      \centering
	   \includegraphics[width=1\textwidth]{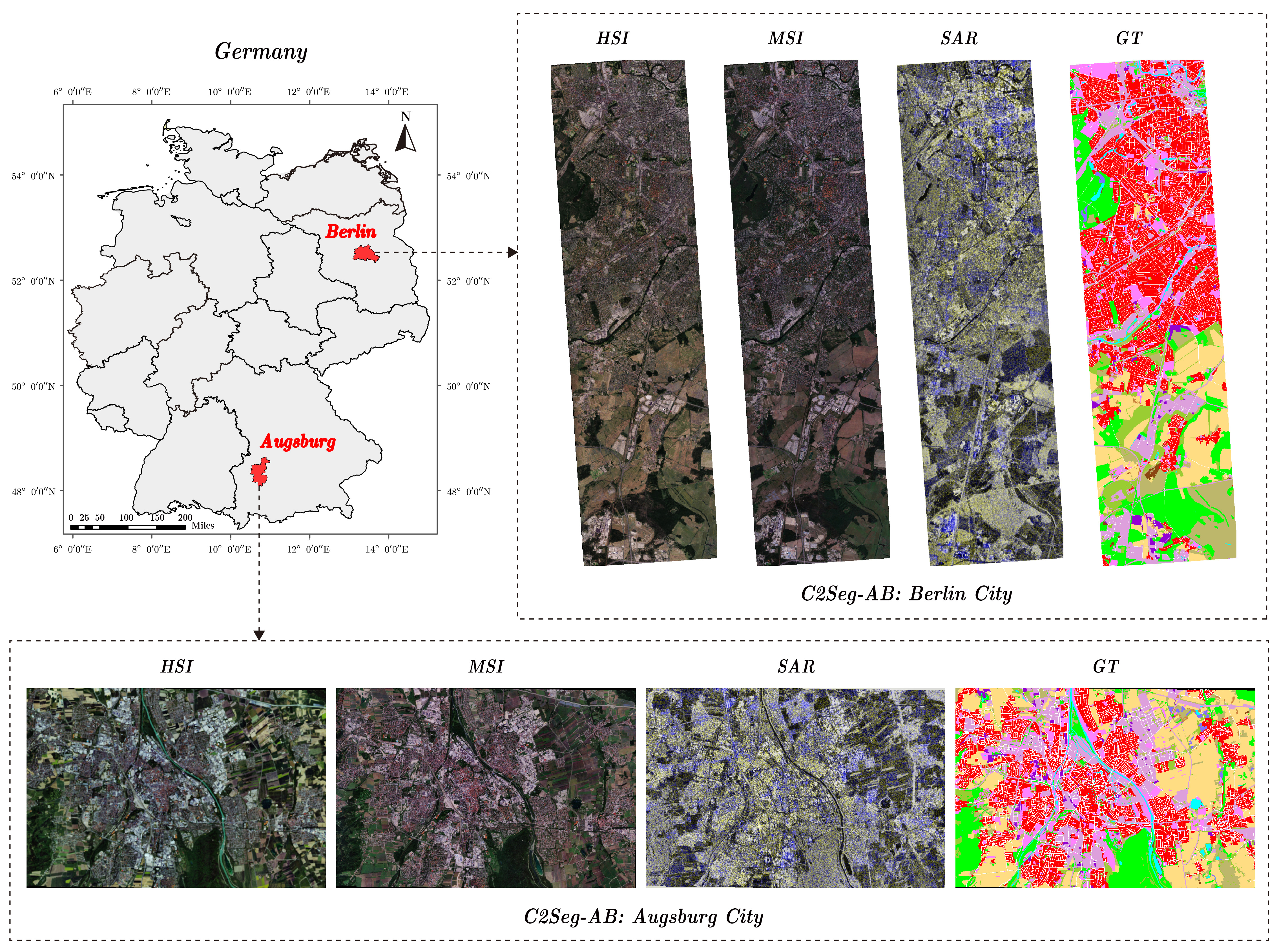}
      \caption{Visualizing C2Seg-AB datasets for semantic segmentation study scene across Berlin and Augsburg cities in Germany using multimodal RS data.}
\label{fig:AB}
\end{figure*}

\subsection{C2Seg-AB}
In C2Seg-AB, the multimodal RS data and labeled semantic categories are prepared across Berlin and Augsburg cities in Germany. C2Seg-AB consists of hyperspectral data from EnMAP, multispectral data from Sentinel-2, and SAR data from Sentinel-1. Fig. \ref{fig:AB} visualizes the C2Seg-AB datasets in terms of scene location, image region, and different modalities with ground truth (GT) of semantic segmentation. 

\textbf{1) EnMAP Hyperspectral Data.} Before launching the EnMAP satellite, the simulation is the main and widely-used way that obtains the EnMAP-related product, which is synthesized by using the full-chain automatic simulation tool, i.e., EeteS \cite{segl2012eetes}, on the high-resolution HyMap or HySpex hyperspectral images. The airborne hyperspectral imaging sensors, i.e., HyMap and HySpex, are used to acquire hyperspectral images over Berlin and Augsburg cities and their neighboring areas. Using EeteS, the corresponding EnMAP images can be simulated by HyMap and HySpex at a ground sample distance (GSD) of 30m, which are openly available form \url{http://doi.org/10.5880/enmap.2016.002} and \url{https://mediatum.ub.tum.de/1657312}, respectively. Further, the two hyperspectral images are upsampled to 10m GSD to keep the identically spatial resolution of all multimodal RS images in the same studied scene. Therefore, the resulting images consist of $2465\times 811$ pixels (Berlin) and $886\times 1360$ pixels (Augsburg), respectively, and they share the same spectral bands (i.e., 242) in the wavelength range of 400nm to 2500nm. More details can be found in \cite{okujeni2016berlin} and \cite{hu2023mdas}.

\textbf{2) Sentinel-2 Multispectral Data.} The Sentinel-2 mission is composed of two twin-orbit satellites (i.e., Sentinel-2A/B) with a combined revisiting time of approximately five days at the equator, the spatial, spectral, and temporal resolution, therefore, makes Sentinel-2 well-suited for dynamic land cover mapping and monitoring. The Sentinel-2 multispectral sensor covers a total of 13 spectral bands ranging from 10m to 60m with different spatial resolutions, and the captured spectral reflectance ranges from visible to NIR and SWIR wavelengths. The best pixels in Sentinel-2 multispectral composite are used in this work, which has been further processed by the SEPAL cloud platform data processing system (sepal.io) of the Food and Agriculture Organization of the United Nations (FAO). Furthermore, the Top of Atmosphere (TOA) reflectance was converted to surface reflectance, and the best pixels were selected from the past three years as of April 2020 using a medoid compositing function, where the radiative transfer models are applied in \cite{hansen2008method} and were later adapted to Sentinel-2 by FAO. In our case, 4 spectral bands are selected from Sentinel-2, e.g., red, green, blue, and near-infrared (NIR), at a GSD of 10m by following a geographic reference of WGS84/UTM Zone 32N. 

\textbf{3) Sentinel-1 SAR Data.} The SAR component is acquired by the Sentinel-1 mission, which is a level-1 Ground Range Detected product obtained by the Interferometric Wide Swath mode. The SAR data is characterized by dual-polarized information with VV and VH channels. The SNAP toolbox is specially designed by the European Space Agency (ESA) for pre-processing Sentinel-1 data to obtain an analysis-ready SAR image, which can be available from the link at \url{https://step.esa.int/main/toolboxes/snap/}. The workflow performed in the SNAP toolbox follows several steps, i.e., precise orbit profile, radiometric calibration, deburst, speckle reduction, and terrain correction. Employing the shuttle radar topography mission, the topographic data are generated well. Different from the Sentinel-2 multispectral image, the Sentinel-1 SAR image is not strictly sampled to the GSD of 10m. Accordingly, the SAR image is geo-coded to be 10m GSD via the bilinear interpolation operator. Finally, the SAR images with two channels, i.e., intensities of VV and VH, are aligned with the pixel-wise EnMAP and Sentinel-2 images.

\textbf{4) Ground Truth of Semantic Segmentation.} Herein, we label the GT of semantic segmentation by retrieving land use and land cover (LULC)-labeled data from OpenStreetMap (OSM) LULC platform at \url{https://osmlanduse.org/} and 12 main classes well-defined in OSMLULC are considered in our case. Accordingly, we manually check the labels within the cities of Berlin and Augsburg and also included the major street network from OSM and appended it to the existing 12 classes, which ensures the granularity and accuracy of the final labeled data. By extending those classes defined in \citep{schultz2017open}, we end up with 13 distinct semantic segmentation features, including urban, industrial, mine, artificial vegetated, arable land, permanent crops, pastures, forests, shrubs, open spaces, inland wetlands, water bodies, and street networks. The elaborately produced LULC maps as GT data (i.e., for the purpose of the semantic segmentation task) in our studied areas are visualized in color (see Fig. \ref{fig:AB}).

\begin{figure*}[!t]
      \centering
	   \includegraphics[width=1\textwidth]{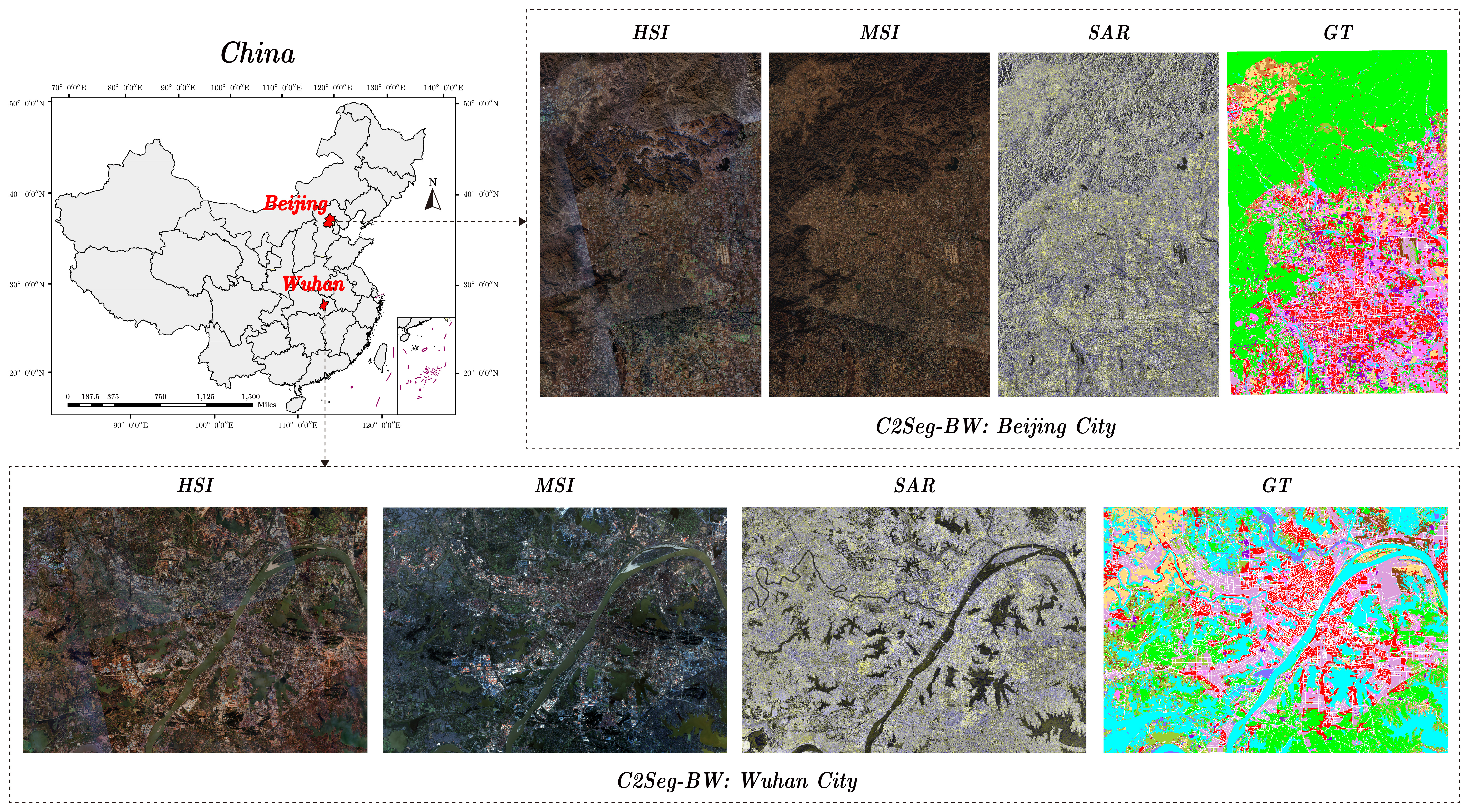}
      \caption{Visualizing C2Seg-BW datasets for semantic segmentation study scene across Beijing and Wuhan cities in China using multimodal RS data.}
\label{fig:BW}
\end{figure*}

\subsection{C2Seg-BW}
The C2Seg-BW dataset provides multimodal RS data and labeled semantic categories across Beijing and Wuhan cities in China, as shown in Fig. \ref{fig:BW}. Similarly, hyperspectral, multispectral, and SAR data are involved in the dataset, which is collected from Gaofen series satellites, such as Gaofen-5, Gaofen-6, and Gaofen-3, respectively. The acquisition dates or satellite perigee passing time of these modality data are late 2019 and early 2020, which ensures that the ground elements remain unchanged as much as possible.

\textbf{1) Gaofen-5 Hyperspectral Data.} The Gaofen-5 hyperspectral data is the level-1A product collected by the Advanced Hyperspectral Imager (AHSI) \cite{liu2019advanced} from the China Center for Resource Satellite Data and Applications (CRESDA). The spatial resolution of the hyperspectral image is around 30m with a narrow swath width of approximately 60km, and there are 330 spectral bands ranging from 400nm to 2500nm. The spectral resolution in the visible and near-infrared (VNIR) region (i.e., 400nm to 1000nm) is about 5nm, while that in the short-wave infrared (SWIR) region (i.e., 1000nm to 2500nm) is about 10nm. 

The hyperspectral images are pre-processed using the ENVI 5.6 software, whose workflow mainly includes radiometric calibration, Fast Line-of-sight Atmospheric Analysis of Spectral Hypercubes (FLAASH) correction, orthorectification, and bands selection. The band selection operation is utilized to massively remove the water vapor absorption, noisy, and bad bands to maintain the image quality. The selected 116 bands are further processed by using the Savitzky-Golay filter. The resulting hyperspectral images are upsampled from 30m to 10m GSD and they then consist of $13474\times 8706$ pixels in Beijing and $6225\times 8670$ pixels in Wuhan, respectively, with a geographic reference of WGS1984 Web Mercator (Auxiliary Sphere).

\textbf{2) Gaofen-6 Multispectral Data.} The Gaofen-6 product is acquired by the specially-designed camera to collect the panchromatic and multispectral images with spatial resolutions of 2m and 8m simultaneously. The multispectral data are used in this paper and pre-processed on the ENVI platform via the standardized processing flow similar to hyperspectral data. To maintain the consistency of the spatial resolution, the four spectral bands in the multispectral image are resampled to 10m. 

\textbf{3) Gaofen-3 SAR Data.} The Gaofen-3 product is collected under the Wide Fine Stripmap mode, yielding a spatial resolution of 10m with a swath width of 100km. The SAR data are prepared by utilizing the functions of de-speckle and terrain correction in the ENVI SARscape Analytics toolbox. The Refined Lee filter \cite{yommy2015sar} with a sliding window size of $5\times 5$ pixels is selected to remove the speckle noises, and the SAR data are corrected employing global digital elevation model (DEM) data in GMTED2010 \cite{danielson2011global}. Similar to Sentinel-1, we adopt the dual-Pol SAR image with HH and HV channels for two studied scenes, and the image size and resolution are the same as those of Ganfen-6 multispectral data.

\begin{figure*}[!t]
	\centering
	\includegraphics[width=1\textwidth]{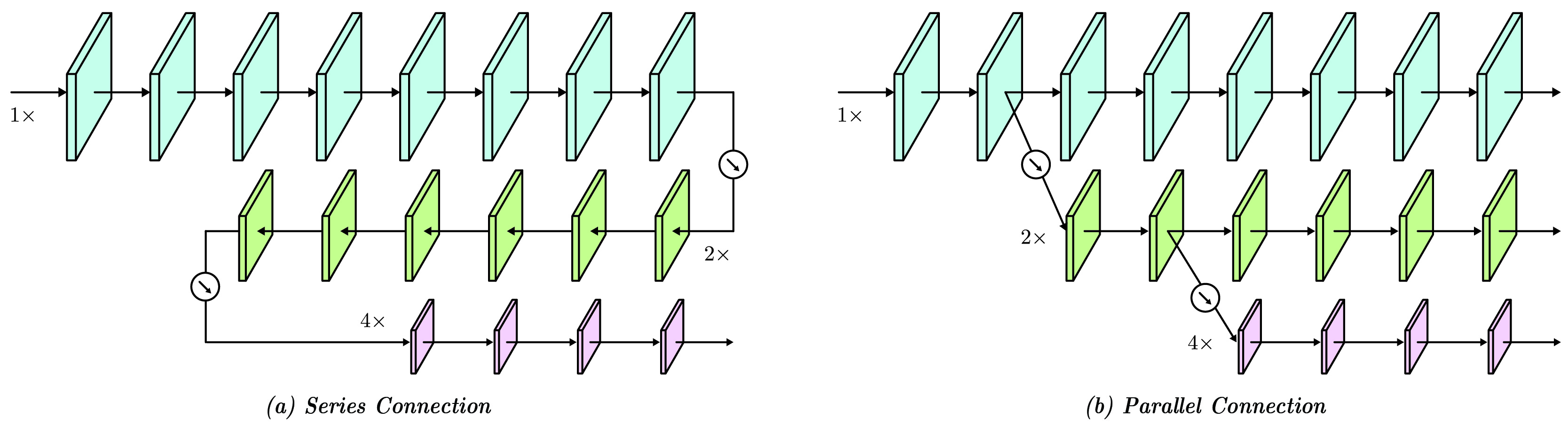}
         \caption{Visualizing the comparison for connection modes of feature maps with different resolutions: (a) Series Connection and (b) Parallel Connection.}
\label{fig:connection}
\end{figure*}

\textbf{4) Ground Truth of Semantic Segmentation.} Similar to C2Seg-AB, we retrieve LULC-labeled data and major street networks within the cities of Wuhan and Beijing (in China) from OSMLULC and OSM, respectively. Herein, we again classify LUCL-labeled data by following the class schema defined in \citep{schultz2017open}, which is based on the widely-accepted Corine Land Cover (CLC) schema \cite{feranec2016european}. However, the availability of OSM data in China is insufficient for semantic labeling. For this reason, we manually map and complete the LULC features by taking multispectral and hyperspectral images as the reference, making it consistent with the labeling schema used in C2Seg-AB datasets. The labeled data of 13 distinct classes serve as a piece of ground-truth information for the following quantitative analysis of cross-city semantic segmentation tasks throughout this paper.

\section{HighDAN: High-Resolution Domain Adaptation Network}
\subsection{A Brief Recall of HR-Net}

Convolutional neural networks (CNNs) have been proven to be effective in learning rich representations from images. Many well-known CNNs-based deep network architectures have been put forward successively, such as AlexNet \cite{krizhevsky2017imagenet}, VGGNet \cite{simonyan2015very}, and GoogleNet \cite{szegedy2015going}. However, there is a potentially common problem in these backbones, i.e., the resolution of the generated feature maps is relatively low when performing the feature extraction by adopting the convolution connection from high resolution to low resolution in series. This inevitably leads to the loss of spatial information. As a result, the traditional solution to this issue is designing an encoder-decoder architecture, i.e., reducing image resolution via the encoder and restoring to high-resolution representations via the decoder. These networks, e.g., U-Net \cite{ronneberger2015u}, SegNet \cite{badrinarayanan2017segnet}, DeconvNet \cite{noh2015learning}, Hourglass \cite{newell2016stacked}, belong to the member of the encoder-decoder structure in essence. Nevertheless, this kind of deep network architecture tends to generate blurred low-resolution feature maps due to multiple convolution operations. These feature maps with different resolutions are further integrated into series connections, raising the risk of the loss of edge details and texture information.

\begin{figure*}[!t]
      \centering
	   \includegraphics[width=0.9\textwidth]{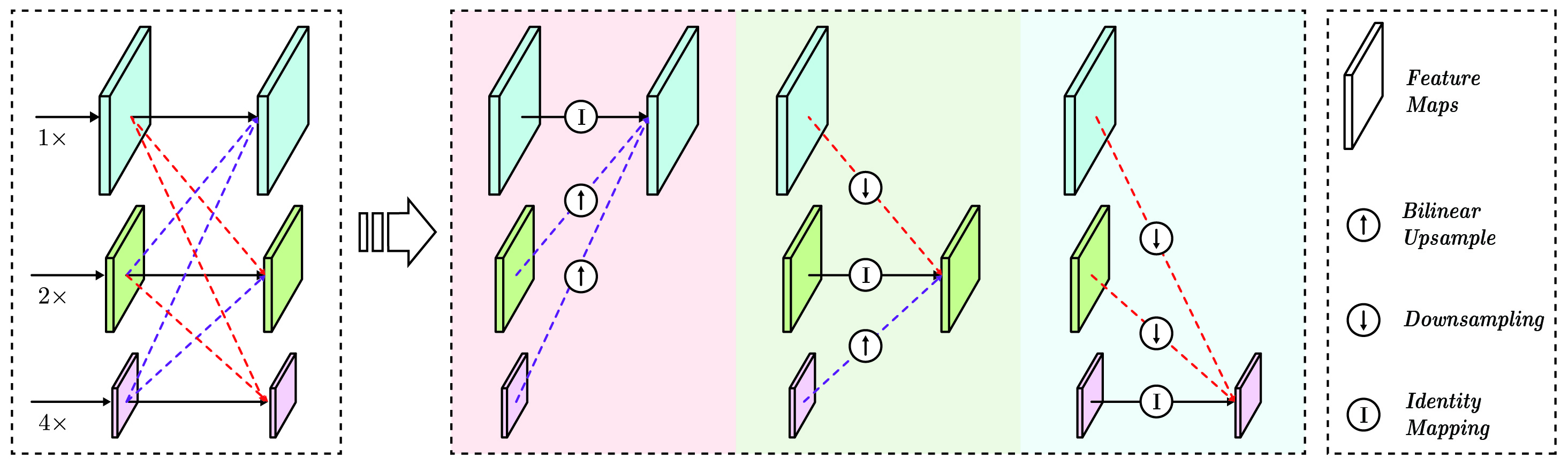}
      \caption{The fusion strategy in HR-Net for feature maps between different resolutions, including the same resolution fusion, upsampling fusion, and downsampling fusion.}
\label{fig:fusion}
\end{figure*}

To overcome the difficulty mentioned above, HR-Net \cite{sun2019deep} is proposed to generate and maintain high-resolution representations. The HR-Net's increments lie in three-folds as follows. 
\begin{itemize}
    \item To connect the high-to-low-resolution convolution streams in a parallel fashion instead of previous series connections, as shown in Fig. \ref{fig:connection} to visualize their differences.
    \item To keep high-resolution representations throughout the whole network architecture.
    \item To exchange the information of feature maps across different resolutions, enabling the compact fusion between high- and low-resolutions to enhance the model's performance. The fusion strategy mainly consists of 1) identity mapping for feature maps with the same resolutions; 2) bilinear upsampling plus $1\times 1$ convolution for feature maps from low to high-resolutions; 3) $3\times 3$ stride convolution for feature maps from high to low-resolutions. Fig. \ref{fig:fusion} illustrates the fusion mode in HR-Net.
\end{itemize}

\begin{figure*}[!t]
      \centering
	   \includegraphics[width=1\textwidth]{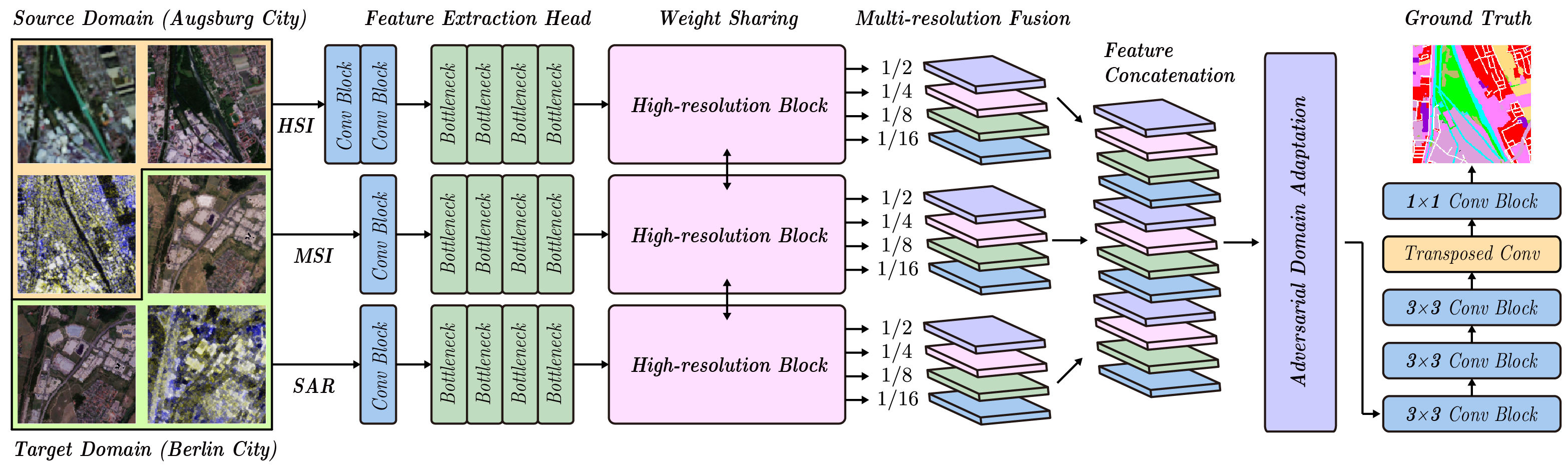}
      \caption{An illustrative workflow of the proposed HighDAN for cross-city semantic segmentation, which mainly consists of feature extraction head, high-resolution (HR) module, multi-resolution fusion, adversarial domain adaptation, and segmentation head (convolution decoder module).}
\label{fig:workflow}
\end{figure*}

\subsection{Method Overview of HighDAN}
Owing to the advancement and superiority of the HR-Net architecture in terms of learning high-resolution representations from images, we propose a novel multimodal HR-Net backbone (i.e., HighDAN) with unsupervised domain adaptation for the cross-city semantic segmentation task using multimodal RS data. Overall, the HighDAN architecture consists of the multimodal encoder, adversarial domain adaptation, and convolution decoder. The design of the domain adaptation module aims to bridge the gap between the representations of source and target domains in an adversarial learning fashion, thereby fully mining the invariant semantic features from multimodal RS data and transferring them across domains. Embedding Dice loss \cite{li2020dice} into networks, HighDAN is capable of weakening the class imbalance effects that tend to be generated in the case of cross-city image interpretation, e.g., semantic segmentation. An illustrative workflow for HighDAN is given in Fig. \ref{fig:workflow}. 

\begin{figure*}[!t]
      \centering
	   \includegraphics[width=1\textwidth]{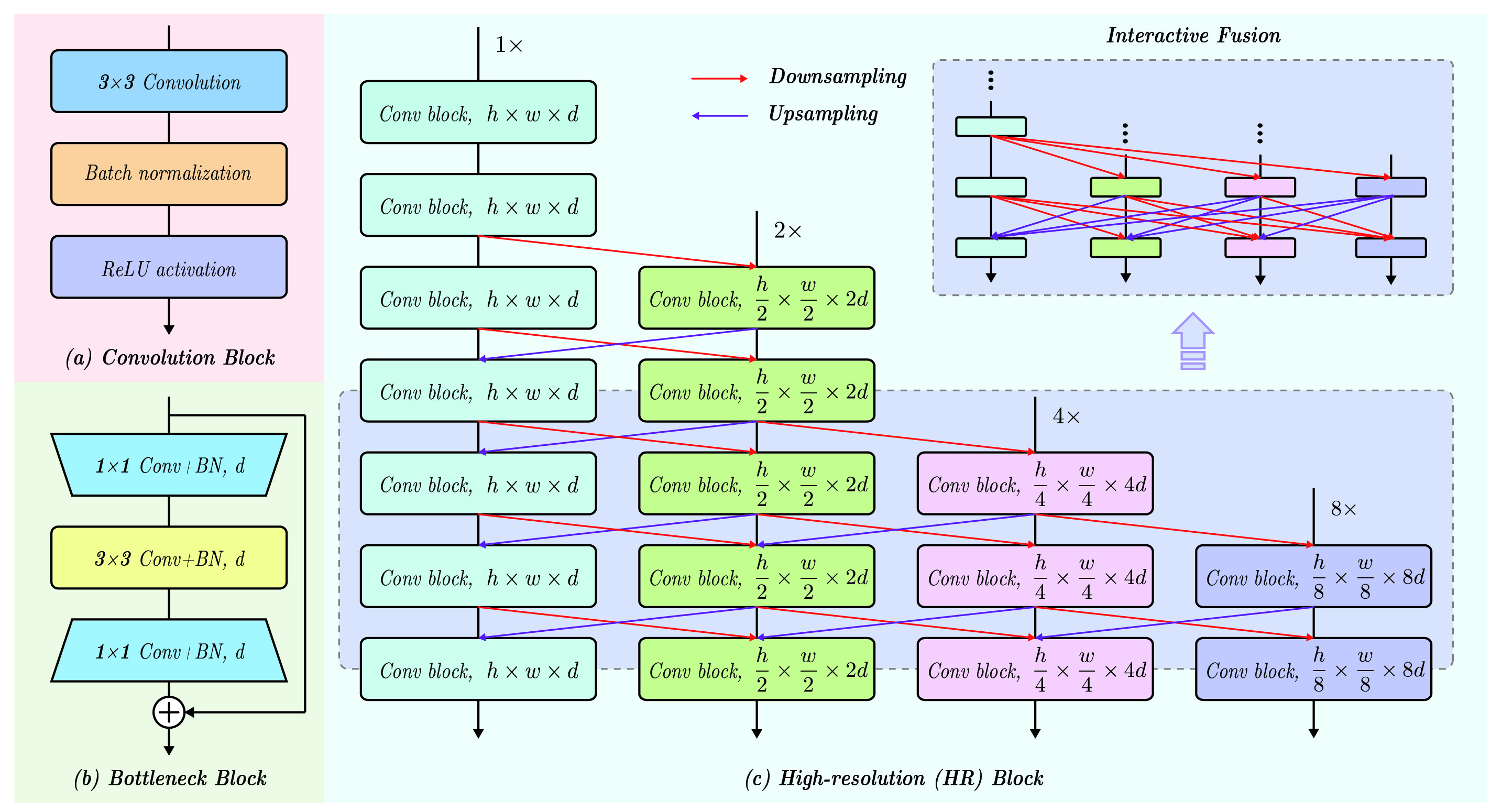}
      \caption{Clarifying details in the multimodal encoder of HighDAN of (a) convolution block, (b) bottleneck block, and (c) HR block, where (a) and (b) form the feature extraction head and (c) is the main body of multimodal HR subnetwork.}
\label{fig:encoder}
\end{figure*}

\subsection{Multimodal Encoder}
The multimodal encoder consists of a feature extraction head and a multimodal high-resolution (HR) subnetwork. As the name suggests, the feature extraction head learns the preliminary representations for different RS modalities by transformations. The head is comprised of the $3\times 3$ convolution block and four bottleneck blocks. Fig. \ref{fig:encoder} visualizes the feature extraction head: (a) convolution block and (b) bottleneck block. The former convolution block can be formulated as
\begin{equation}
\label{eq1}
\begin{aligned}
       \bm{Z}_{k}= f_{\bm{W}_{k},\bm{B}_{k}}(\bm{X}_{k}),
\end{aligned}
\end{equation} 
where $k$ is the index (e.g., $1,2,...$) for different RS modalities, and $\bm{X}$ and $\bm{Z}$ denote the input modality image and the feature representations via the convolution block, respectively. The function $f(\cdot)$, i.e., the convolution block, is unfolded as $3\times 3$ convolution operation, batch normalization (BN), and ReLU activation function, which is with respect to the network variables of weights $\bm{W}$ and biases $\bm{B}$. {Given that hyperspectral data typically possesses a significantly higher dimensionality compared to multispectral and SAR data, it is common practice to employ dimensionality reduction techniques (e.g., PCA) to preprocess the data before feeding it into networks. Additionally, to ensure compatibility with the input dimensions of bottleneck blocks, several extra convolutional layers are utilized for all input data, facilitating seamless integration within the network architecture.} The later bottleneck block is expressed by
\begin{equation}
\label{eq2}
\begin{aligned}
       \bm{Q}_{k}=g_{\bm{W}_{k},\bm{B}_{k}}(\bm{Z}_{k}),
\end{aligned}
\end{equation}
where $\bm{Q}$ denotes the feature representations via the bottleneck block. The bottleneck block can be represented as the function $g(\cdot)$ with respect to the to-be-learned network variables: $\bm{W}$ and $\bm{B}$, which can be unfolded as $1\times 1$ convolution, BN, $3\times 3$ convolution, BN, $1\times 1$ convolution, and BN in sequence. {To provide a further explanation, the multimodal encoder in HighDAN initiates with a three-stream network architecture that takes as input multimodal RS data, including hyperspectral, multispectral, and SAR (see Fig. \ref{fig:workflow}). This architecture is instrumental in elucidating the approach used to effectively combine data from diverse RS modalities.}

The multimodal HR subnetwork well inherits attributes of HR-Net that can extract HR image representations. {Following the HR-Net, the input RS modality image is firstly downsampled by convolution operations with a 2-stride as the main stem.} By gradually adding high-to-low-resolution streams, feature maps with different resolutions are then connected and fused in parallel to acquire diversified resolution representations. The process can be written as 
\begin{equation}
\label{eq3}
\begin{aligned}
       \bm{V}_{k}=h_{\bm{W}_{k},\bm{B}_{k}}(\bm{Q}_{k}),
\end{aligned}
\end{equation}
where $\bm{V}_{k}$ denotes the HR representations of the $k$-th modality via the multimodal HR subnetwork. The function $h(\cdot)$ is defined as the multimodal HR subnetwork by copying the HR module in HR-Net \cite{sun2019deep}, which is illustrated in Fig. \ref{fig:encoder} (c) with HR block. That is, it consists of a multi-resolution group convolution and a multi-scale fusion layer. The former refers to a regular convolution for each resolution stream over different spatial resolutions separately, and the latter aims to perform an interactive fusion of feature maps across scales. It should be noted that the HR module for different RS modalities is shared in terms of network parameters to capture the high-quality multimodal characteristics more steadily. The outputs from each resolution stream are re-scaled to the same resolution as HR representations through bilinear upsampling, achieving the multi-resolution fusion via feature stacking.

\begin{figure*}[!t]
      \centering
	   \includegraphics[width=1\textwidth]{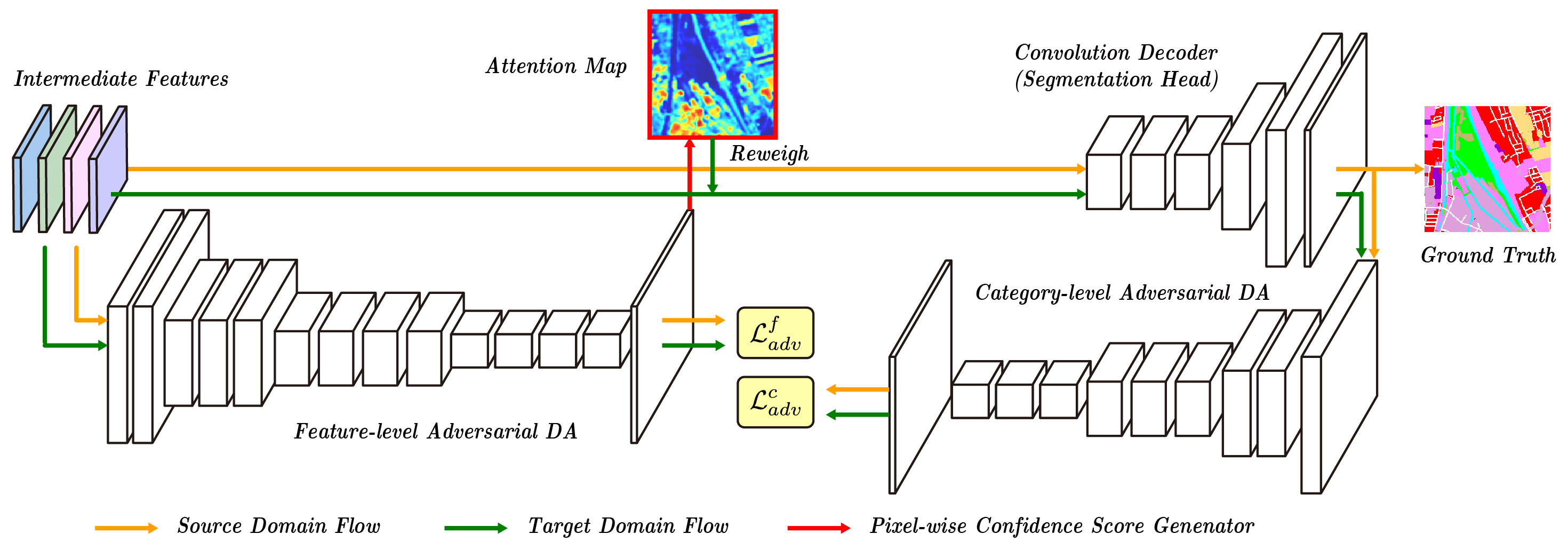}
      \caption{A diagram of adversarial domain adaptation used in HighDAN, which consists of feature-level adversarial DA and category-level adversarial DA. The pixel-wise attention map is generated by the feature-level adversarial DA and explored for reweighing and correcting the feature representations from the target domain, making it compatible with features from the source domain. The category-level adversarial DA can further refine the segmentation results.}
\label{fig:DA}
\end{figure*}

\subsection{Adversarial Domain Adaptation}
According to the adversarial learning in GAN, the image-to-image translation techniques \cite{isola2017image} enable the pixel-level alignment and knowledge conversion between source and target domains. This further provides possible and potential solutions to the cross-domain semantic segmentation task. Prior to conducting DA, it is essential to concatenate all feature maps obtained from the various multimodal streams, denoted as $\bm{V}=\{\bm{V}_{k}\}_{k=1}^{m}$. This consolidation of feature maps is a crucial step in the process. Inspired by \cite{yu2021dast}, we adopt two types of DA modules based on the adversarial learning strategy to align representations of source and target domains at both feature-level and category-level. On the one hand, the feature-level DA module attempts to reduce biases of cross-domain intermediate feature maps (i.e., $\bm{V}$) obtained from the multimodal HR encoder. Herein, pixel-wise confidence scores that can reflect the degree of local alignment in different domains are generated from the discriminator, which can be used to reweigh the intermediate features $\bm{V}$ to correct the representation shift between different domains locally. This yields the aligned representations as $\bm{A}$. On the other hand, the category-level DA module aims to enhance global semantic alignment from the label distribution perspective, which is used in the final prediction phase. The global semantic alignment operation can be regarded as a kind of soft constraint on the category centers, which drives the same category closer to each other in different domains. Visually, Fig. \ref{fig:DA} gives the corresponding diagram of adversarial DA used in HighDAN.

\subsection{Convolution Decoder}
Given the aligned feature representations $\bm{A}$ via DA, a segmentation head in the form of the convolution decoder is further applied on $\bm{A}$ to progressively reconstruct feature maps consistent with the size of semantic labels, which can be formulated by
\begin{equation}
\label{eq4}
\begin{aligned}
       \bm{U}=T_{\bm{W},\bm{B}}(\bm{A}),
\end{aligned}
\end{equation}
where $\bm{U}$ denotes the predicted semantic label map, and the function $T(\cdot)$ represents the decoder module that consists of convolution, BN, ReLU activation function, and 2x upsampling operation. 

\begin{algorithm}[!t]
\label{alg1}
\caption{A flowchart of the proposed HighDAN}
\KwIn{Different RS modality data ($\bm{X}_{k}$) including labeled (source domain, $\bm{X}^{s}_{k}$) and unlabeled (target domain, $\bm{X}^{t}_{k}$) samples, and semantic segmentation labels ($\bm{Y}$) corresponding to labeled samples.}
\KwOut{Model, predicted segmentation results $\hat{\bm{Y}}$}
 \textbf{Step 1 (Data preparation):} Band-wise normalization for $\bm{X}_{k}$, and feed the normalized data into networks with corresponding labels.\\
 \textbf{Step 2 (Model Training):}\\
  \For{$t=1$ to $T$}{
      {Part 1: Multimodal Encoder}\\
      \quad {\small 1) Feature Extraction Head:\\
      \quad \qquad Convolution block: Feature representations $\bm{Z}_{k}$ obtained by Eq. (\ref{eq1});\\
      \quad \qquad Bottleneck block: Feature representations $\bm{Q}_{k}$ obtained by Eq. (\ref{eq2});}\\
      \quad {\small 2) Multimodal HR Head: HR representations $\bm{V}_{k}$ obtained by Eq. (\ref{eq3}).}\\
      {Part 2: Adversarial Domain Adaptation}\\
      \quad {\small 1) Feature Concatenation: $\bm{V}=\{\bm{V}_{k}\}_{k=1}^{m}$;\\}
      \quad {\small 2) Feature Alignment: Aligned representations $\bm{A}$ via feature-level DA on $\bm{V}$.\\}
      {Part 3: Convolution Decoder}\\
      \quad {\small 1) Predict semantic labels $\bm{U}$ by Eq. (\ref{eq4});}\\
      \quad {\small 2) Compute the overall loss by Eq. (\ref{eq5});}\\
      \quad {\small 3) Gradient backpropagation and update networks.}\\
   }
  \textbf{Step 3 (Model Inference):}\\
  \quad {\small 1) Normalize multimodal testing data in a band-wise fashion;}\\
  \quad {\small 2) Feed the normalized data into learned networks;}
  \\
  \quad {\small 3) Obtain the predicted labels and output the final segmentation results.}
\end{algorithm}

\subsection{Model Training}
{A flowchart illustrating the proposed HighDAN model is outlined in \textbf{Algorithm 1}, with step-by-step procedures provided for clarity.} Let $\bm{X}\in \mathbb{R}^{hw\times N}$ and $\bm{Y}\in \mathbb{R}^{l\times N}$ be the input images and the ground truth (GT) of semantic segmentation labels with $hw$ and $l$ dimensions, respectively, by $N$ pixels. Then, $\bm{x}_{i}$ and $\bm{y}_{i}$ are denoted to be the corresponding $i$-th element  (or pixel). With these definitions, the network concerning the to-be-updated parameters of $\bm{W}$ and $\bm{B}$ is trained by optimizing the following objective function. The overall loss $\mathcal{L}$ in objective function is
\begin{equation}
\label{eq5}
\begin{aligned}                \mathcal{L}=\mathcal{L}_{seg}+\lambda\mathcal{L}^{f}_{adv}+\mu\mathcal{L}^{c}_{adv},
\end{aligned}
\end{equation}
where $\lambda$ and $\mu$ are defined as the penalty parameters to balance different terms in the training phase, and we set them to be both 0.5 empirically and experimentally. More specifically, the three terms are detailed in the following.

The first term in Eq. (\ref{eq5}) is the segmentation loss, which consists of multi-class cross-entropy loss and Dice loss, i.e.,
\begin{equation}
\label{eq6}
\begin{aligned}         
       \mathcal{L}_{seg}=\mathcal{L}_{MCE}+\mathcal{L}_{Dice}.
\end{aligned}
\end{equation}
$\mathcal{L}_{MCE}$ calculates the loss for each pixel equally, and $\mathcal{L}_{Dice}$ can alleviate the negative effects due to the imbalanced training samples, e.g.,
\begin{equation}
\label{eq7}
\begin{aligned}         
       \mathcal{L}_{Dice}=1-\frac{2\sum_{i=1}^{N}\bm{y}_{i}\bm{\hat{y}}_{i}}{\sum_{i=1}^{N}\bm{y}_{i}+\sum_{i=1}^{N}\bm{\hat{y}}_{i}},
\end{aligned}
\end{equation}
where $\bm{\hat{y}}_{i}$ denotes the predicted semantic label in the $i$-th pixel. 

The second term in Eq. (\ref{eq5}) is the feature-level adversarial loss. Unlike the vanilla GAN that utilizes the classic cross-entropy loss to train the discriminator, the least square loss in \cite{mao2017least} is exploited in our DA task to avoid the gradient vanishing issue. Suppose the input modality data $\bm{X}^{s}$ is from the source domain and $\bm{X}^{t}$ is from the target domain, the generator $E_{f}$ and discriminator $D_{f}$ can be alternatively optimized by minimizing
\begin{equation}
\label{eq8}
\begin{aligned}         
       &\mathcal{L}^{f}_{adv}(D_{f})=\mathbb{E}_{\bm{X}^{s}}[(D_{f}(\bm{V}^{s})-0)^{2}]+\mathbb{E}_{\bm{X}^{t}}[(D_{f}(\bm{V}^{t})-1)^{2}],\\
       &\mathcal{L}^{f}_{adv}(E_{f})=\mathbb{E}_{\bm{X}^{t}}[(D_{f}(E_{f}(\bm{X}^{t}))-0)^{2}],
\end{aligned}
\end{equation}
where $\bm{V}^{s}$ and $\bm{V}^{t}$ are the feature maps (e.g., using Eq. (\ref{eq3})) extracted from the source domain and target domain, respectively, via multimodal HR encoder module (collectively known as the generator $E_{f}$ in our case). To ensure the stability of feature maps of the target domain, we optimize $\bm{V}^{t}$ by using the updated rule of $\bm{V}^{t}_{new}=\bm{V}^{t}+\bm{V}^{t}\odot\alpha$, where $\alpha$ denotes the attention map.

The third term in Eq. (\ref{eq5}) is the category-level adversarial loss. The analogy to the second term, the adversary is performed at the category level to improve the global adaptation ability in networks. We thus have the following adversarial loss:
\begin{equation}
\label{eq9}
\begin{aligned}         
       &\mathcal{L}^{c}_{adv}(D_{c})=\mathbb{E}_{\bm{X}^{s}}[(D_{c}(\bm{U}^{s})-0)^{2}]+\mathbb{E}_{\bm{X}^{t}}[(D_{c}(\bm{U}^{t})-1)^{2}],\\
       &\mathcal{L}^{c}_{adv}(P_{c})=\mathbb{E}_{\bm{X}^{t}}[(D_{c}(P_{c}(\bm{X}^{t}))-0)^{2}],
\end{aligned}
\end{equation}
where $\bm{U}^{s}$ and $\bm{U}^{t}$ are the output's decoder maps (e.g., using Eq. (\ref{eq4})) of the source domain and target domain via the proposed HighDAN, that is $P_{c}$ as well.

\section{Experiments}
\subsection{Experimental Preparation}

\subsubsection{Implementation Details}
 The proposed HighDAN is implemented on the PyTorch platform, and all deep models are trained using CPU with i7-6850K, RAM with 128GB, and GPU with 11GB NVIDIA GTX1080Ti. The Adam \cite{kingma2014adam} is selected as the network optimizer with the iterations of 6000 epochs for C2Seg-AB and 10000 epochs for C2Seg-BW, respectively. The learning rates of the segmentation network and discriminator are both 0.0001 with a batch size of 16. By cropping the whole scene images with the sliding window at certain intervals, we collect 273 (or 7140) and 140 (or 850) images with the size of $128\times 128$ (or $256\times 256$) as a source domain for training and as a target domain for testing, respectively, on C2Seg-AB (or C2Seg-BW) datasets. 

\begin{table*}[!t]
\centering
\caption{Layer-wise network configuration of the proposed HighDAN. Conv, BN, HR, and Num are abbreviations of convolution, batch normalization, high resolution, and number, respectively.}
\vspace{2mm}
\resizebox{0.92\textwidth}{!}{ 
\begin{tabular}{c||ccc|c}
\toprule[1.5pt]
 & Hyperspectral & Multispectral & SAR & Output Dimension\\
 \hline \hline
\multirow{6}{*}{Convolution Block} & $3\times 3$ Conv & $3\times 3$ Conv & $3\times 3$ Conv  & \multirow{6}{*}{64}\\
& BN & BN & BN &\\
& ReLU & ReLU & ReLU &\\
& $3\times 3$ Conv  & -- & -- &\\
& BN & -- & -- &\\
& ReLU & -- & -- &\\
\hline
Bottleneck Module & Bottleneck Block$\ast$4 & Bottleneck Block$\ast$4 & Bottleneck Block$\ast$4 & 48\\
\hline
\multirow{2}{*}{Feature Encoding 1} & Basic HR Block$\ast$4 & Basic HR Block$\ast$4 & Basic HR Block$\ast$4 & \multirow{2}{*}{48/96}\\
& Sum Fusion & Sum Fusion & Sum Fusion & \\
\hline
\multirow{2}{*}{Feature Encoding 2} & Basic HR Block$\ast$4 & Basic HR Block$\ast$4 & Basic HR Block$\ast$4 & \multirow{2}{*}{48/96/192}\\
& Sum Fusion & Sum Fusion & Sum Fusion & \\
\hline
\multirow{2}{*}{Feature Encoding 3} & Basic HR Block$\ast$4 & Basic HR Block$\ast$4 & Basic HR Block$\ast$4 & \multirow{2}{*}{48/96/192/384}\\
& Sum Fusion & Sum Fusion & Sum Fusion & \\
\hline
Fusion Layer & \multicolumn{3}{c|}{Feature Concatenation} & 720$\ast$3\\
\hline
Adversarial Module & \multicolumn{3}{c|}{Output the attention scores to reweigh feature maps} & 720$\ast$3\\
\hline
\multirow{16}{*}{Convolution Decoder Module} & \multicolumn{3}{c|}{$3\times 3$ Conv} & \multirow{3}{*}{256}\\
& \multicolumn{3}{c|}{BN} & \\
& \multicolumn{3}{c|}{ReLU} & \\
\cline{2-5}
& \multicolumn{3}{c|}{$3\times 3$ Conv} & \multirow{3}{*}{128}\\
& \multicolumn{3}{c|}{BN} & \\
& \multicolumn{3}{c|}{ReLU} & \\
\cline{2-5}
& \multicolumn{3}{c|}{$3\times 3$ Conv} & \multirow{3}{*}{64}\\
& \multicolumn{3}{c|}{BN} & \\
& \multicolumn{3}{c|}{ReLU} & \\
\cline{2-5}
& \multicolumn{3}{c|}{Transposed Convolution} & \multirow{2}{*}{Num of Class}\\
& \multicolumn{3}{c|}{$1\times 1$ Conv} & \\
\bottomrule[1.5pt]
\end{tabular}}
\label{tab:network_configuration}
\end{table*}

\subsubsection{Network Configuration}
To enable the reconstruction of the proposed semantic segmentation network, we particularize the HighDAN architecture layer by layer. HighDAN successively starts with convolution blocks, and four bottleneck blocks are connected. Behind it, three feature encoding modules are adopted, each consisting of four basic HR blocks. The convolution decoder module is finally added with the combination of four decoding blocks. Between the two modules, an adversarial block and a concatenation-based fusion layer are embedded. For more details, the layer-wise network configuration of HighDAN is listed in Table \ref{tab:network_configuration}.

\subsubsection{Evaluation Metrics}
We evaluate the cross-city semantic segmentation performance qualitatively and quantitatively in terms of three metrics in common use: overall accuracy (OA), mean intersection over union (mIoU), and mean F1 score (mF1). OA, also known as pixel accuracy (PA), collects each pixel prediction:
\begin{equation}
\label{eq10}
\begin{aligned}
      OA=\frac{\sum_{i=1}^{l}p_{ii}}{\sum_{i=1}^{l}\sum_{j=1}^{l}p_{ij}},
\end{aligned}
\end{equation}
where $i$, $j$, and $l$ represent the real value, predicted value, and the total number of classes, respectively, and the $p_{ij}$ denotes the number of pixels that predict the $i$-th class as the $j$-th class. mIoU computes the intersection and union of two sets, which is defined by
\begin{equation}
\label{eq11}
\begin{aligned}
      mIoU=\frac{1}{l}\sum_{i=1}^{l}\frac{p_{ii}}{\sum_{j=1}^{l}p_{ij}+\sum_{j=1}^{l}p_{ji}-p_{ii}}.
\end{aligned}
\end{equation}
mF1 score is the harmonic mean of precision ($P$) and recall (R), which is given by
\begin{equation}
\label{eq12}
\begin{aligned}
      mF1=\frac{1}{l} \times \frac{2\times P\times R}{P+R},
\end{aligned}
\end{equation}
where 
\begin{equation}
\label{eq13}
\begin{aligned}
      P=\sum_{i=1}^{l}\frac{p_{ii}}{\sum_{j=1}^{l}p_{ij}+p_{ii}},\;\;\;
      R=\sum_{i=1}^{l}\frac{p_{ii}}{\sum_{j=1}^{l}p_{ji}+p_{ii}}.
\end{aligned}
\end{equation}

\begin{table*}[!t]
\centering
\caption{Quantitative performance comparison of deep semantic segmentation networks {in terms of OA, mIoU, mF1, and F1 scores for each class as well as the model's computational complexity (FLOPs) and parameters} on C2Seg-AB datasets. The symbol `--' denotes no pixels correctly identified. The best result is marked in bold.}
\vspace{2mm}
\resizebox{1\textwidth}{!}{ 
\begin{tabular}{l||ccccccc|c}
\toprule[1.5pt]
Class Name & DeepLab & SegNet & FastFCN & AdaptSeg & DSAN & DualHR & SegFormer & HighDAN \\
\hline\hline
Surface water & 0.31 & 18.53 & 24.24 & 22.96 & 12.20 & 38.49 & 37.45 & \bf 52.06\\
Street network & 9.54 & 3.00 & 13.00 & 13.61 & 5.80 & 20.98 & 0.79 & \bf 21.35\\
Urban fabric & 56.58 & 58.08 & 60.75 & 66.00 & 71.66 & 62.73 & 69.84 & \bf 72.52\\
Industrial, commercial, and transport & 35.86 & 39.88 & 41.48 & 42.99 & 26.15 & 48.13 & 57.98 & \bf 62.37\\
Mine, dump, and construction sites	& -- & 5.72 & 6.30 & 6.90 & 15.96 & 17.72 & \bf 22.32 & 21.95\\
Artificial vegetated areas & 25.35 & 0.33 & 16.21 & 19.11 & 24.38 & 18.01 & 26.56 & \bf 33.41\\
Arable land	& 53.28 & 58.34 & 48.01 & 48.47 & 55.85 & 61.14 & 53.91 & \bf 65.92\\
Permanent crops & -- & -- & -- & -- & -- & -- & -- & \bf 1.17\\
Pastures & 1.69 & 34.91 & 8.02 & 0.83 & 21.49 & 0.23 & 35.97 & \bf 37.87\\
Forests	& 62.85 & 34.20 & 51.47 & 53.43 & 65.72 & 53.52 & 70.87 & \bf 74.35\\
Shrub & 0.59 & 1.60 & 0.39 & 11.23 & \bf 17.64 & 2.34 & 5.06 & 14.51\\
Open spaces with no vegetation & -- & -- & -- & -- & -- & -- & -- &  --\\
Inland wetlands & -- & -- & -- & -- & -- & -- & -- & --\\
\hline\hline
OA (\%) & 42.46 & 43.40 & 43.51 & 44.53 & 47.68 & 48.05 & 53.40 & \bf 57.66\\
mIoU (\%) & 12.64 & 12.65 & 13.33 & 14.30 & 16.33 & 16.49 & 20.16 & \bf 24.76\\
mF1	(\%) & 18.93 & 19.59 & 20.76 & 21.96 & 24.37 & 24.87 & 29.29 & \bf 35.19\\
\hline\hline
{FLOPs (B) / GFLOPs} & {56.42} & {30.53} & {38.58} & {38.34} & {53.67} & {37.46} & {\bf 5.82} & {40.11}\\
{Params (M)} & {72.68} & {29.48} & {56.70} & {44.86} & {107.44} & {\bf 15.33} & {28.97} & {16.55}\\
\bottomrule[1.5pt]
\end{tabular}
}
\label{table:Performance_C2Seg-AB}
\end{table*}

\subsubsection{Comparison with State-of-the-art Models}
We select current state-of-the-art (SOTA) semantic segmentation models for qualitative and quantitative performance comparison using multimodal RS data in the cross-city case. They are DeepLabv3 \cite{chen2017deeplab}, SegNet \cite{badrinarayanan2017segnet}, FastFCN \cite{wu2019fastfcn}, AdaptSeg \cite{tsai2018learning}, deep subdomain adaptation network (DSAN) \cite{zhu2021deep}, Dual-stream HR-Net (DualHR) \cite{ren2022dual}, {SegFormer \cite{xie2021segformer}}, and our proposed HighDAN. {The \cite{chen2017deeplab}, \cite{badrinarayanan2017segnet}, \cite{wu2019fastfcn}, \cite{ren2022dual} and \cite{xie2021segformer} models fail to consider data shifts between different domains, while the rest effectively embed the DA strategy into networks.} It is worth noting that we prioritize using the same network configurations (given in the original literature) for compared approaches. Further, the relevant parameters can be slightly adjusted, making it applicable to the segmentation experiments of multimodal RS data.

\begin{table*}[!t]
\centering
\caption{Quantitative performance comparison of deep semantic segmentation networks {in terms of OA, mIoU, mF1, and F1 scores for each class as well as the model's computational complexity (FLOPs) and parameters} on C2Seg-BW datasets. The symbol `--' denotes no pixels correctly identified. The best result is marked in bold.}
\vspace{2mm}
\resizebox{1\textwidth}{!}{ 
\begin{tabular}{l||ccccccc|c}
\toprule[1.5pt]
Class Name & DeepLab & SegNet & FastFCN & AdaptSeg & DSAN & DualHR & SegFormer & HighDAN \\
\hline\hline
Surface water & 50.42 & 37.87 & 45.39 & 59.57 & -- & 60.51 & \bf 78.49 & 78.37\\
Street network & 16.30 & 6.17 & 2.38 & \bf 17.05 & -- & 0.29 & 0.05 & 0.58\\
Urban fabric & 33.75 & 34.86 & 38.44 & 25.06 & \bf 42.08 & 0.76 & 30.90 & 40.04\\
Industrial, commercial, and transport & 2.48 & 1.87 & 27.63 & 32.77 & 25.68 & 24.19 & 20.38 & \bf 43.67\\
Mine, dump, and construction sites & 1.70 & 1.36 & 0.86 & \bf 1.94 & 1.22 & 0.26 & 2.52 & 1.67\\
Artificial vegetated areas & 8.97 & 2.23 & 8.83 & 8.99 & 9.55 & 4.19 & \bf 10.79 & 9.28\\
Arable land	& -- & 16.23 & -- & 9.11 & \bf 25.37 & -- & 18.01 & 0.43\\
Permanent crops & -- & 1.57 & -- & 0.30 & -- & -- & \bf 1.77 & 0.10\\
Pastures & -- & -- & -- & -- & -- & -- & -- & --\\
Forests	& 1.68 & 13.31 & 4.14 & 32.98 & 46.85 & 32.70 & 38.72 & \bf 47.22\\
Shrub & 0.72 & -- & -- & -- & -- & 0.22 & \bf 10.48 & 0.26\\
Open spaces with no vegetation & -- & -- & -- & -- & -- & \bf 0.33 & 0.01 & --\\
Inland wetlands & -- & -- & -- & -- & -- & \bf 0.01 & \bf 0.01 & --\\
\hline\hline
OA (\%) & 19.51 & 15.94 & 21.22 & 29.26 & 18.55 & 31.97 & 33.56 & \bf 39.58\\
mIoU (\%) & 5.45 & 5.17 & 5.96 & 8.92 & 7.09 & 6.17 & 10.89 & \bf 11.92\\
mF1	(\%) & 8.92 & 8.88 & 9.82 & 14.44 & 11.60 & 9.53 & 16.32 & \bf 17.69\\
\hline\hline
{FLOPs (B) / GFLOPs} & {213.01} & {122.13} & {154.32} & {144.58} & {214.67} & {149.86} & {\bf 23.30} & {160.45}\\
{Params (M)} & {72.68} & {29.48} & {56.70} & {44.86} & {107.44} & {\bf 15.33} & {28.97} & {16.55}\\
\bottomrule[1.5pt]
\end{tabular}
}
\label{table:Performance_C2Seg-BW}
\end{table*}

\subsection{Quantitative Evaluation on C2Seg Datasets}
Tables \ref{table:Performance_C2Seg-AB} and \ref{table:Performance_C2Seg-BW} quantify the cross-city semantic segmentation performance by comparing current SOTA deep models with our HighDAN in terms of pixel-wise OA, mIoU, mF1, and F1 scores for each class as well as {the model's computational complexity (FLOPs) and parameters on C2Seg datasets (C2Seg-AB and C2Seg-BW, respectively).} 

By and large, the cross-city segmentation performance of deep networks without the consideration of data shifts across domains (e.g., DeepLabv3, SegNet) is inferior to that of those models that effectively embed the DA strategy into networks. SegNet shows comparable performance with DeepLabv3 in terms of OA, mIoU, and mF1 on C2Seg-AB Datasets, while SegNet and DeepLabv3 hold similar segmentation accuracies on C2Seg-BW Datasets. For those DA-guided segmentation networks, the adversarial DA methods (e.g., FastFCN, AdaptSeg) show competitive results compared to DSAN based on the local maximum mean discrepancy. Although FastFCN and AdaptSeg perform moderately lower than DSAN at an average decrease of 3\%$\sim$4\% OAs, 2\%$\sim$3\% mIoUs, and 2\%$\sim$4\% mF1s, respectively, yet their F1 scores for each category are holistically comparable to DSANs' and the main differences lie in certain special categories, e.g., \textit{Pastures}, \textit{Forests}, \textit{Shrub}, etc. on the C2Seg-AB datasets. It is important to note that when confronted with more complex and extensive datasets e.g., C2Seg-BW, the generalization capability of DSAN appears to be somewhat constrained in comparison to FastFCN and AdaptSeg.

Furthermore, the HR-Net backbone architecture can offer greater potential for extracting a wealth of semantic information from multimodal RS data in comparison with the CNNs-based backbone in the semantic segmentation task. For example, DualHR brings increments of 13\% OA based on DSAN on C2Seg-BW datasets, but the performance is basically identical to those on C2Seg-AB datasets, compared to DSAN. {However, it is essential to note that transformer-based methods (i.e., SegFormer) consistently demonstrate competitive and stable performance on both C2Seg datasets, achieving the second-highest results across all evaluation indices.} Not unexpectedly, the proposed HighDAN achieves the best segmentation performance by 4.26\%, 4.60\%, and 5.90\% gains in OA, mIoU, and mF1 (\textit{cf.} SegFormer) on C2Seg-AB datasets, while there is also a nearly similar trend, even higher performance (e.g., over 6\% OA increase), on C2Seg-BW datasets. A more noteworthy point to demonstrate the superiority of HighDAN lies in that HighDAN obtains the highest F1 scores in many dominated categories, e.g., \textit{Surface water}, \textit{Street network}, \textit{Urban fabric}, \textit{Arable land}, \textit{Forests}, etc. on either C2Seg-AB or C2Seg-BW datasets. We have to admit, however, that C2Seg is a very challenging semantic segmentation dataset. It is observed that some categories are hardly identified, that is, the segmentation results for certain classes are 0\% and few are approximately close to 0\%.

\begin{figure*}[!t]
       \centering
       \includegraphics[width=1\textwidth]{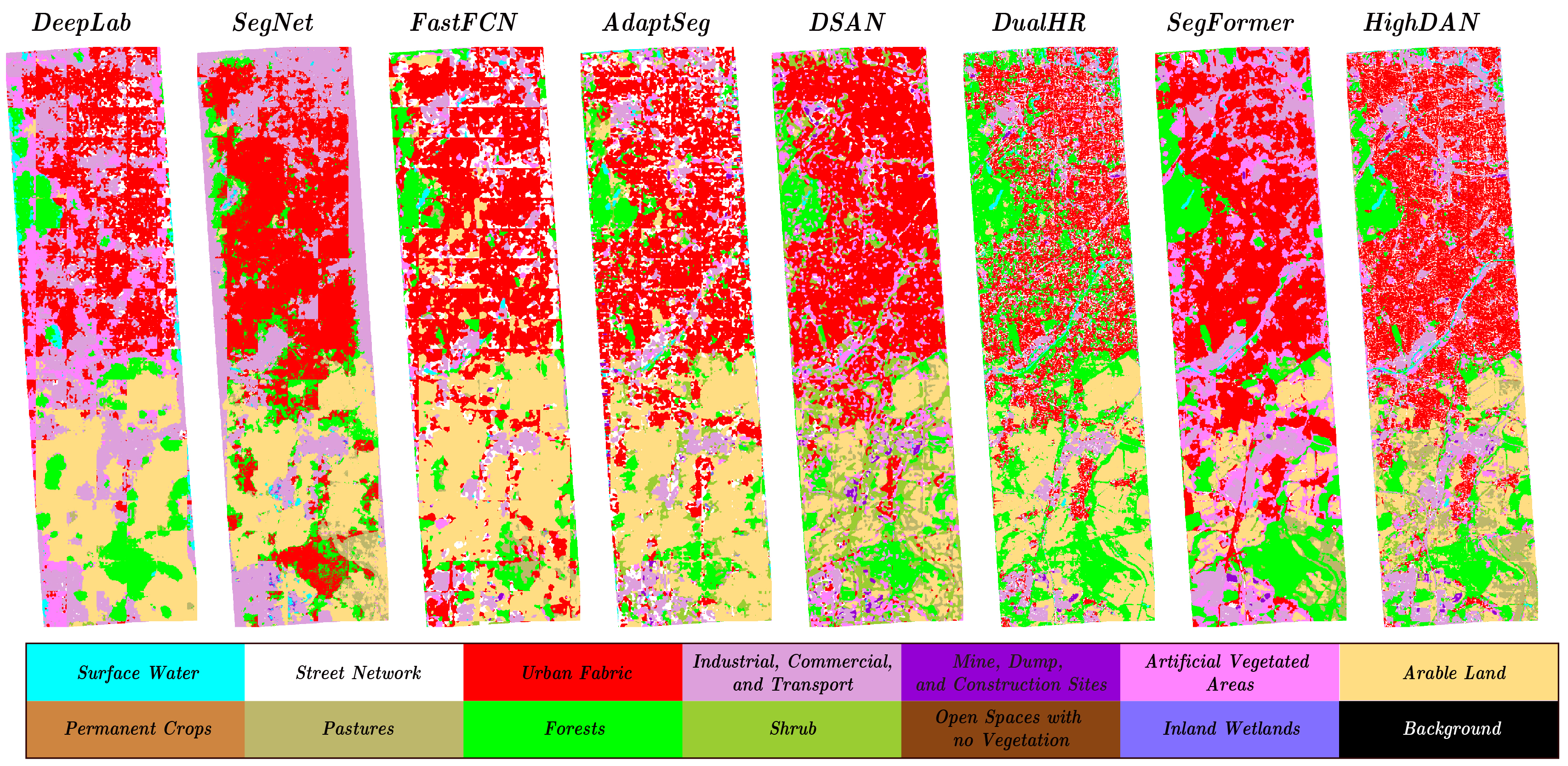}
       \caption{Visualization of semantic segmentation results obtained by successively using DeepLab, SegNet, FastFCN, AdaptSeg, DSAN, DualHR, {SegFormer}, and our proposed HighDAN on C2Seg-AB (testing set: Berlin) datasets.}
\label{fig:CM_Berlin}
\end{figure*}

\begin{figure*}[!t]
       \centering
       \includegraphics[width=1\textwidth]{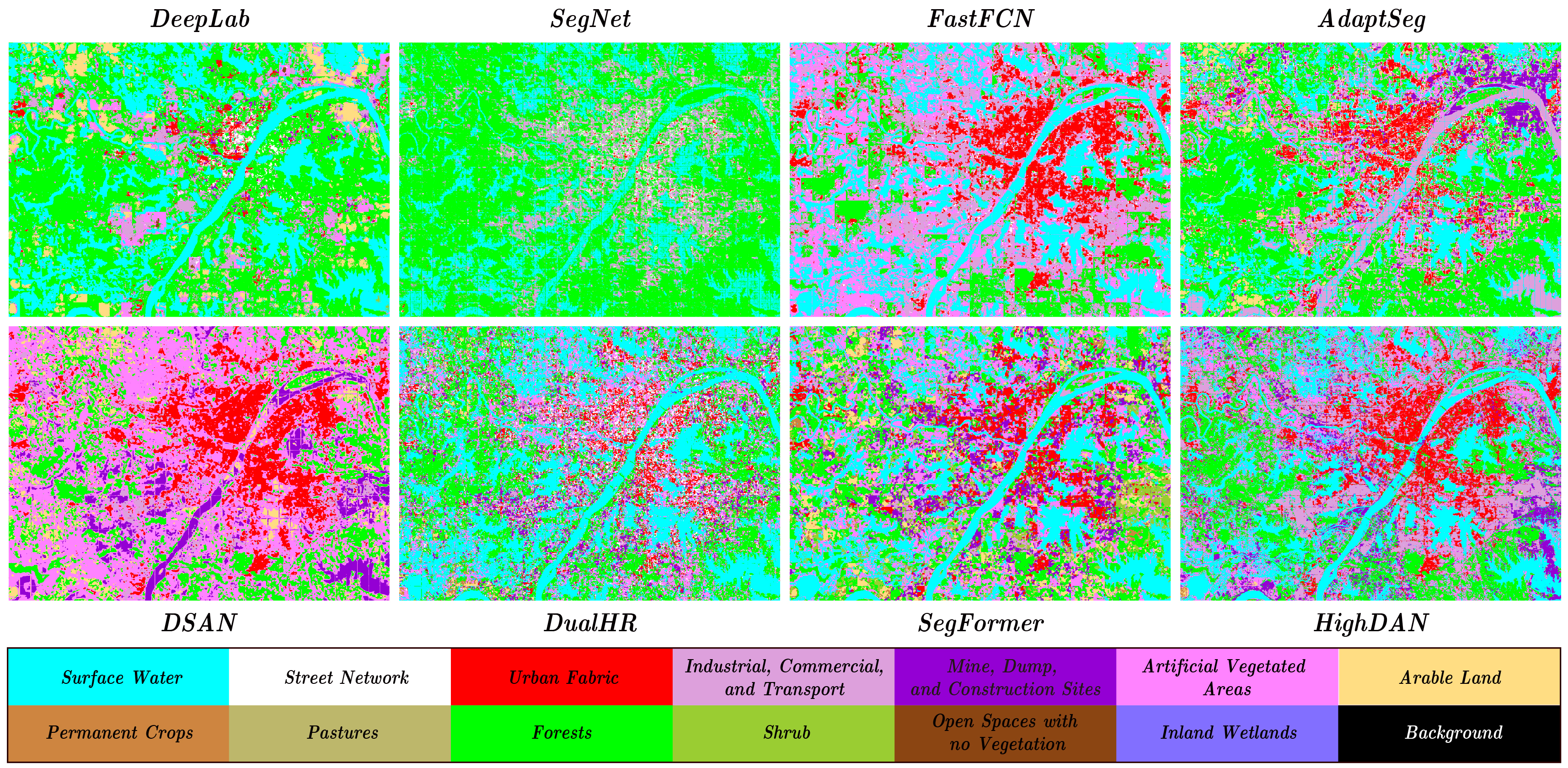}
       \caption{Visualization of semantic segmentation results obtained by successively using DeepLab, SegNet, FastFCN, AdaptSeg, DSAN, DualHR, {SegFormer}, and our proposed HighDAN on C2Seg-BW (testing set: Wuhan) datasets.}
\label{fig:CM_Wuhan}
\end{figure*}

\begin{figure*}[!t]
      \centering
	   \includegraphics[width=1\textwidth]{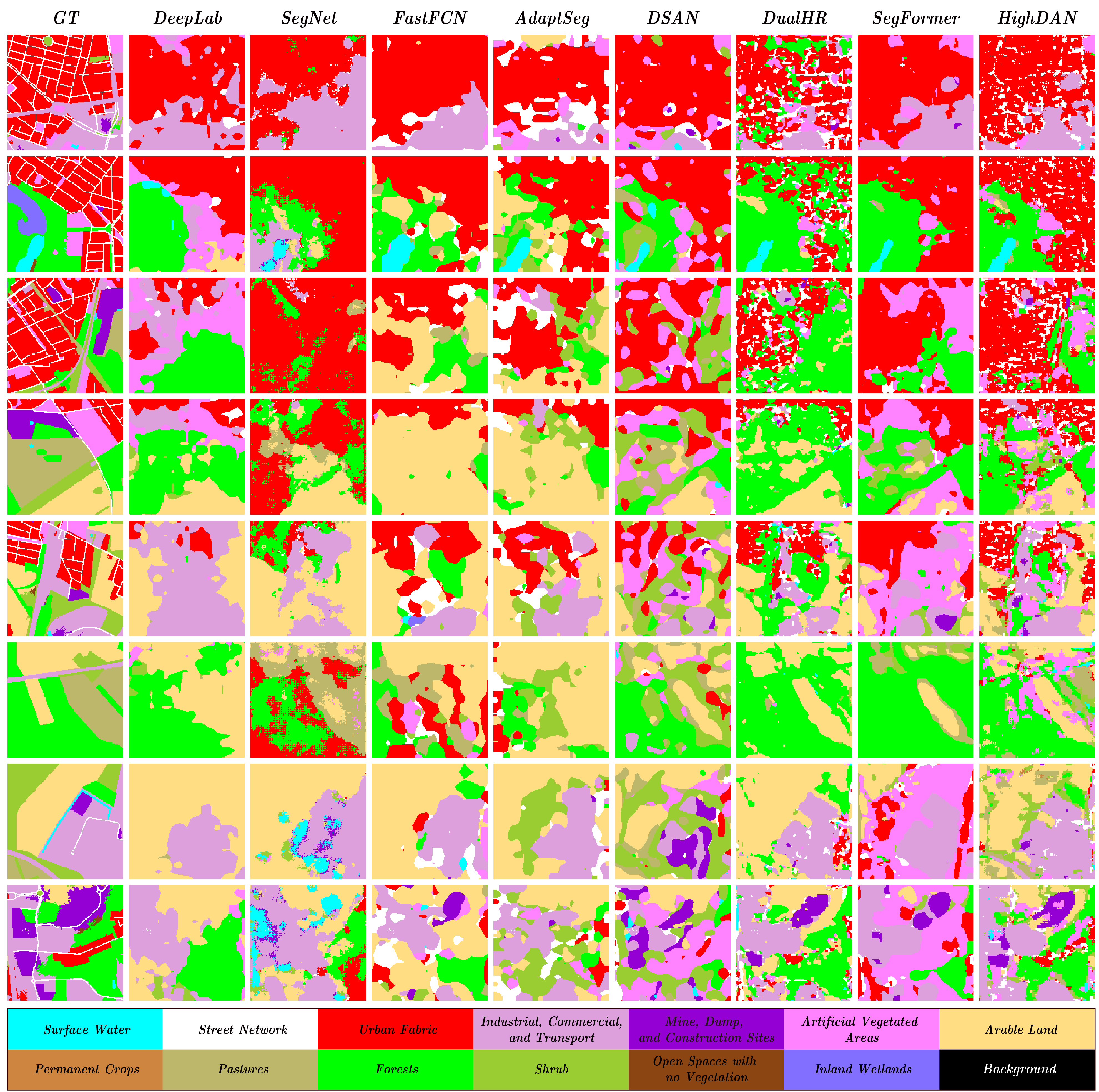}
      \caption{Visualizing semantic segmentation results of sub-regions corresponding to Fig. \ref{fig:CM_Berlin} for all compared models.}
\label{fig:Patch_Berlin}
\end{figure*}

\subsection{Visual Comparison on C2Seg Datasets}
Figs. \ref{fig:CM_Berlin} and \ref{fig:CM_Wuhan} visualize the segmentation maps of eight different algorithms in terms of 13 semantic categories for the whole scenes of Berlin city and Wuhan city on C2Seg datasets. There is a more significant visual difference between predicted segmentation results and GT (on both Berlin and Wuhan scenes) in DeepLab and SegNet. On the one hand, \textit{Pastures} are prone to be wrongly classified as \textit{Arable land}, while \textit{Inland Wetlands} are heavily identified to be \textit{Forests} in the Wuhan scene. On the other hand, \textit{Urban fabric} and \textit{Industrial, commercial, and transport} are easily confused due to their similar spectral characteristics and functions. Compared to the first two methods, FastFCN has visible advantages in discriminating the semantic category of \textit{Urban Fabric} and \textit{Artificial vegetated areas}, while AdaptSeg is capable of identifying \textit{Arable Land} more accurately (despite the over-recognition of \textit{Shrub} and \textit{Pastures} being \textit{Arable Land}). We have to admit, however, that the ability of AdaptSeg to classify urban-related semantic elements remains limited. DSAN is a good recognizer for urban-related and vegetation semantic categories, which can well distinguish \textit{Urban fabric} and \textit{Industrial, commercial, and transport} as well as \textit{Forests} and \textit{Arable land}. In the family of HR-Net, DualHR is sensitive to capturing water bodies from a big urban scene but fails to detect urban accurately, while the proposed HighDAN visually shows, as expected, comparatively realistic segmentation maps closer to GT (\textit{cf.} SegFormer). In particular, water bodies, urban, and forests have nearly identical semantic segmentation profiles to those in GT. There is, notwithstanding, considerable room for improvement in HighDAN, to further enhance the identification and recognition ability in \textit{Arable Land}, \textit{Street Network}, and \textit{Inland Wetlands}. 

\begin{figure*}[!t]
      \centering
	   \includegraphics[width=1\textwidth]{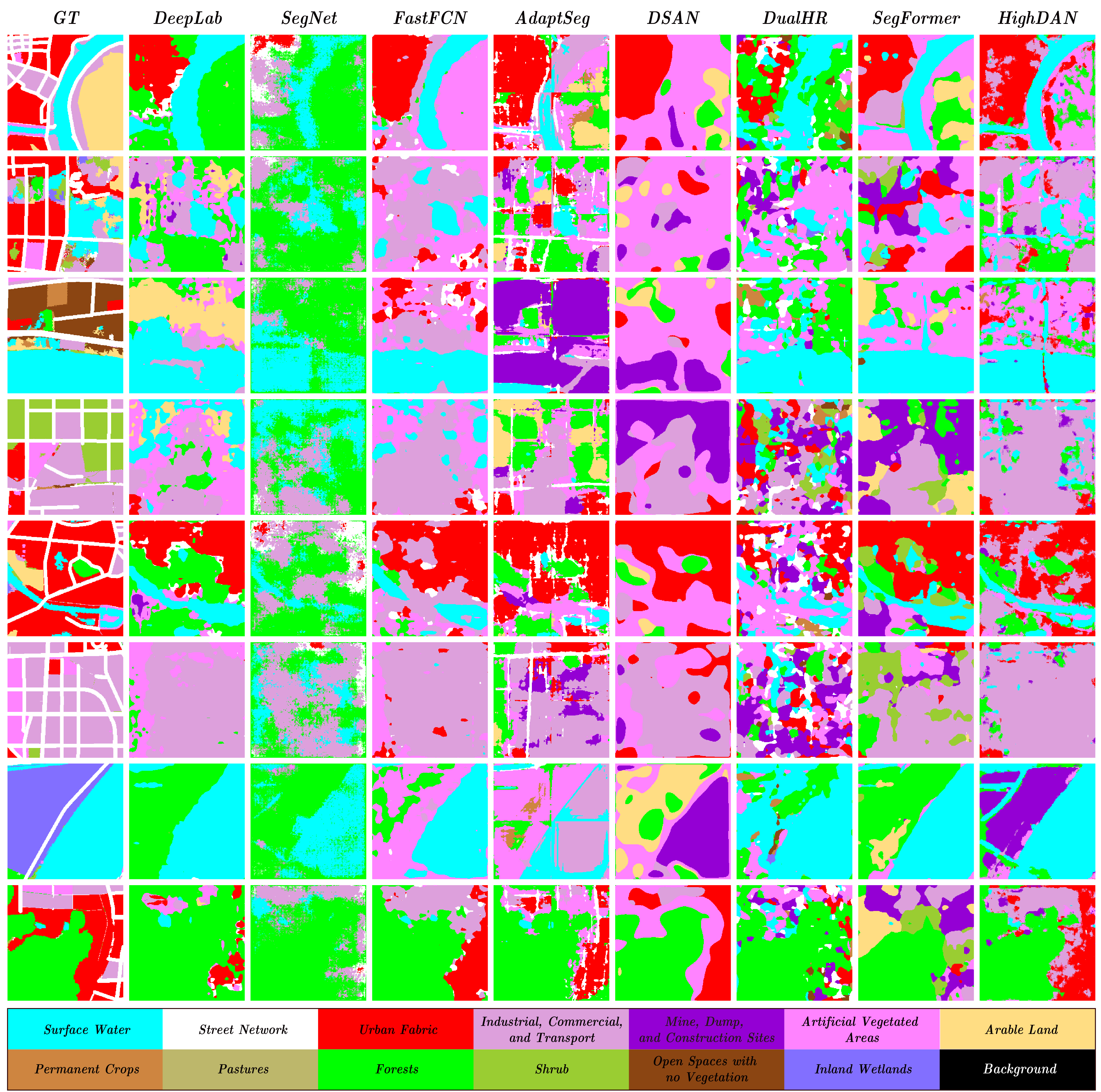}
      \caption{Visualizing semantic segmentation results of sub-regions corresponding to Fig. \ref{fig:CM_Wuhan} for all compared models.}
\label{fig:Patch_Wuhan}
\end{figure*}

\begin{figure*}[!t]
      \centering
	   \includegraphics[width=1\textwidth]{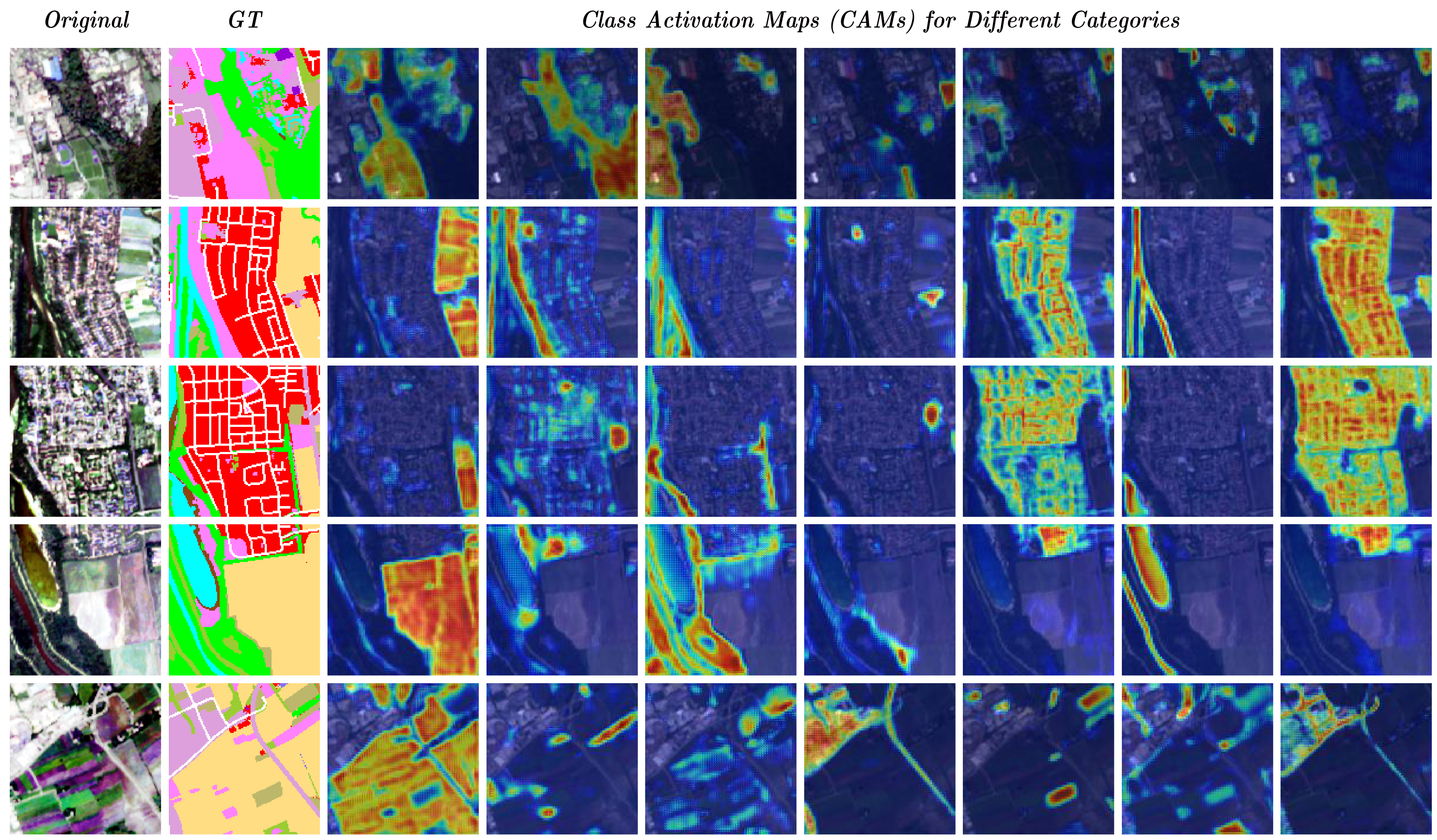}
      \caption{{Visualization of class activation maps (CAMs) using HighDAN on the C2Seg-AB datasets.}}
\label{fig:attention}
\end{figure*}

In addition to the scene-wide segmentation visualization, we also provide detailed segmentation results in sub-regions, as shown in Figs. \ref{fig:Patch_Berlin} and \ref{fig:Patch_Wuhan} corresponding to Figs. \ref{fig:CM_Berlin} and \ref{fig:CM_Wuhan}, respectively. The visual comparison of local semantic segmentation results highlights the advantages of the proposed HighDAN in terms of preserving fine-grained details of objects in RS images. Further, HighDAN is capable of effectively capturing small-scale features and details of the objects. This was particularly evident in the cases of man-made objects with intricate shapes and textures, where HR-Net-based models (i.e., DualHR, HighDAN) are apt to segment the objects without losing important details. In comparison, the baseline methods, such as DeepLab and SegNet, yield segmentation results with a severe loss of detailed information, which shows their limitations in capturing tiny and irregular objects or structures. While other compared methods have demonstrated some improvement in identifying semantic categories with varying shapes, their ability in recognition accuracy and boundary segmentation remains limited. Yet the visual analysis also reveals that our HighDAN can effectively adapt to changes in imaging conditions and variability in object appearance across domains or cities, resulting in improved segmentation accuracy and robustness. It should be noted, however, that some categories are almost entirely misclassified in certain sub-images, such as \textit{Artificial vegetated areas}, \textit{Open spaces with no vegetation}, \textit{Inland wetlands}, \textit{Shrub}. To sum up, these observations highlight the need for continued exploration and optimization of semantic segmentation methods in the aspects of HR feature extraction and DA enhancement. 

{To further assess the effectiveness of our proposed HighDAN model in extracting class-related semantic information, we visualize class activation maps (CAMs) \cite{zhou2016learning} on the C2Seg-AB datasets, as shown in Fig. \ref{fig:attention}. These visualizations demonstrate that HighDAN excels in capturing high-level semantic information with precise class activation, even for small classes, e.g., \textit{Street Network}. This capability underscores the model's proficiency in semantic segmentation tasks.}

\begin{table*}
\centering
\caption{Ablation analysis on C2Seg-AB datasets, where `Bottleneck' and `HR' represent the bottleneck feature extraction module and the high-resolution module, while `Feature DA' and `Category DA' denote feature-level domain adaptation module and category-level domain adaptation module, respectively. The best results are marked in bold.}
\vspace{2mm}
\resizebox{1\textwidth}{!}{ 
\begin{tabular}{l|ccccc|ccc}
\toprule[1.5pt]
Model & Bottleneck & HR & Feature DA & Category DA & Dice Loss & OA (\%) & mIoU (\%) & mF1 (\%)\\
\hline\hline
SegNet & \xmark & \xmark & \xmark & \xmark & \xmark & 43.40 & 12.65 & 19.59\\
HR-Net & \xmark & \cmark & \xmark & \xmark & \xmark & 48.53 & 15.86 & 23.35\\
HR-Net & \cmark & \cmark & \xmark & \xmark & \xmark & 49.85 & 18.95 & 27.80\\
HR-Net & \cmark & \cmark & \xmark & \xmark & \cmark & 50.59 & 19.90 & 29.04\\
\hline
HighDAN & \cmark & \cmark & \xmark & \cmark & \cmark & 53.74 & 20.62 & 29.84\\
HighDAN & \cmark & \cmark & \cmark & \xmark & \cmark & 52.73 & 21.51 & 31.09\\
HighDAN & \cmark & \cmark & \cmark & \cmark & \xmark & 56.44 & 24.03 & 34.32\\
\hline\hline
HighDAN & \cmark & \cmark & \cmark & \cmark & \cmark & \bf 57.66 & \bf 24.76 & \bf 35.19\\
\bottomrule[1.5pt]
\end{tabular}}
\label{table:Ablation}
\end{table*}

\subsection{Ablation Study}
The proposed HighDAN takes the multimodal HR-Net as the network backbone, which consists of several key modules, such as multimodal HR encoder (Bottleneck + HR), DA (Feature-level DA + Category-level), and Dice loss. To evaluate the importance of these modules for cross-city semantic segmentation using multimodal RS data, we implement the ablation study on C2Seg-AB datasets. Table \ref{table:Ablation} details the performance gain by combining different components in terms of OA, mIoU, and mF1. 

SegNet follows the classic encoder-decoder backbone and serves as the baseline (without any advanced components involved), yielding relatively poor segmentation performance. By integrating the bottleneck and the advanced HR feature extractor, HR-Net significantly improves at an increment of 6.45\% OA, 6.30\% mIoU, and 8.21\% mF1. With the Dice loss, HR-Net considers the class imbalance issue and shows competitive results, but without DA, it inevitably meets the performance bottleneck in the cross-city task. The adversarial DA strategy bridges the gap across domains effectively from feature-level and category-level perspectives. HighDAN demonstrated a noteworthy improvement in OA, with a substantial 6\% enhancement over HR-Net, and exhibited remarkable increases of approximately 5\% in the pivotal semantic segmentation metrics, i.e., mIoU and mF1. Notably, balancing samples of different categories via Dice loss also plays a prominent role in HighDAN. As can be seen from Table \ref{table:Ablation}, HighDAN with Dice loss can further improve the cross-city semantic segmentation performance by at least 1.2\% OA based on that without the loss. {We also present results that facilitate a comparison between scenarios involving HS data and those without HS data in terms of OA, mIoU, and mF1: (57.66\%, 35.19\%, 24.76\%) \textit{vs.} (53.91\%, 31.81\%, 21.74\%).}

\begin{figure*}[!t]
      \centering
	   \includegraphics[width=1\textwidth]{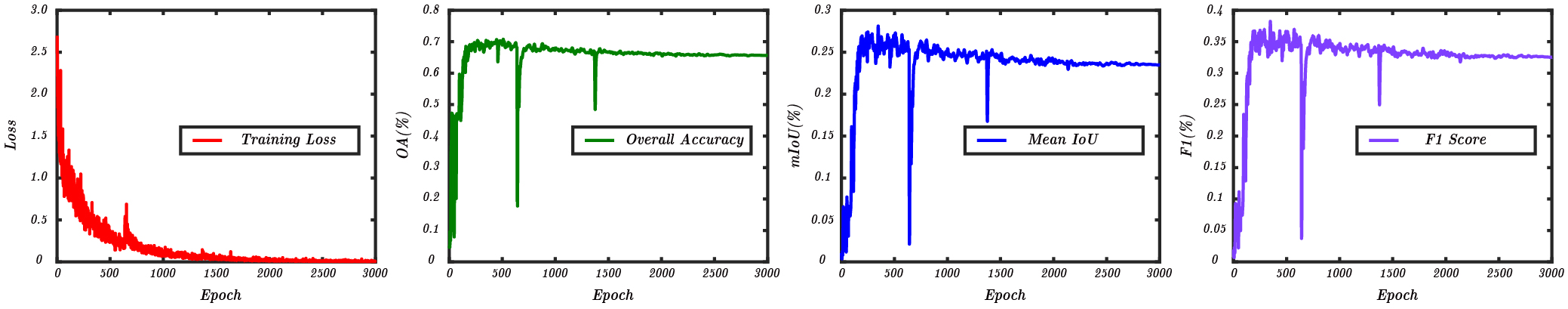}
      \caption{{Performance analysis of the proposed HighDAN in robustness in terms of training loss, OA, mIoU, and mF1 score on the individual C2Seg-AB datasets.}}
\label{fig:robustness}
\end{figure*}

{In addition, we presented the results, which included training loss, OA, mIoU, and mF1, for individual datasets using an 8:2 training and testing ratio, specifically focusing on C2Seg-AB. This comprehensive evaluation process allowed us to assess the performance and robustness of the proposed HighDAN model. Fig. \ref{fig:robustness} illustrates that the training loss of the model exhibits a consistent decrease throughout the training process, indicating the model's stability and robust convergence during learning. As expected, there is a similar trend in segmentation performance (i.e., OA, mIoU, mF1) across individual C2Seg-AB datasets.}

\section{Conclusion}
Fast monitoring and understanding of urban environments are inseparable from explosively developing RS techniques. The success of RS enables the accurate identification and detection of materials of interest in complex urban scenes. As a primary and indispensable research topic, the semantic segmentation of RS images has long dominated the overwhelming role in the land use land cover classification of urban environments. However, these well-designed and dedicated segmentation methodologies are, for the most part, applicable only to one single city case. This severely hinders the application deployments across cities or regions, since urban planning and management, e.g., policy-making, land use, spatial layout, information transfer, etc., have to accommodate multi-city studies. 

For the reason mentioned above, we in this paper focus on investigating cross-city semantic segmentation and provide solutions accordingly. The solutions are two-fold. On the one hand, we build a multimodal RS benchmark dataset (i.e., C2Seg) to solve the issue of insufficient discriminative information by only using single modality RS data for cross-city semantic segmentation. On the other hand, we propose a cutting-edge deep network architecture, HighDAN for short, by embedding the adversarial learning-based DA's idea into HR-Net with Dice Loss (to reduce the effects of the class imbalance), making it largely possible to break the semantic segmentation performance bottleneck in terms of accuracy and generalization ability from cross-city studies. Extensive experiments conducted on the C2Seg datasets demonstrate that our HighDAN achieves the best segmentation performance, which beats other SOTA competitors in almost all important indices. Moreover, we will also release the C2Seg benchmark datasets and the corresponding source codes, contributing to the interpretation research of urban environments across cities. 

In future work, we aim to extend the C2Seg datasets in a wide range of cities on a national scale and even a global scale for the better study of cross-city semantic segmentation. In particular, the development of hyperspectral RS, especially concerning its application on a large scale, is indeed an issue that warrants urgent attention and exploration, due to certain inherent imaging constraints associated with hyperspectral RS technology. Furthermore, more advanced AI models should be developed and made accessible by further considering explicit and explainable knowledge embedding, e.g., geometric priors, climate characteristics, and urban morphological properties, to guide deep networks to learn more accurate segments and promote the model's generalization ability across cities.

\section*{Acknowledgements} The authors would like to thank Ms. Zhu Han and Ms. Luyang Cai for pre-processing the Gaofen data used in this paper. This work was supported by the National Key Research and Development Program of China under Grant 2022YFB3903401, the National Natural Science Foundation of China under Grant 42241109 and Grant 42271350, the MIAI@Grenoble Alpes (ANR-19-P3IA-0003), the Klaus Tschira Stiftung (KTS) Heidelberg, and the AXA Research Fund.
\clearpage
\bibliography{reference}

\end{document}